\documentclass[Afour,sageh,times]{sagej}

\usepackage{moreverb,url}

\usepackage[colorlinks,bookmarksopen,bookmarksnumbered,citecolor=red,urlcolor=red]{hyperref}

\usepackage{xspace}
\usepackage{algorithm, algpseudocode}
\usepackage{cleveref}
\usepackage{array}
\usepackage{adjustbox}
\usepackage[normalem]{ulem}

\newcommand\BibTeX{{\rmfamily B\kern-.05em \textsc{i\kern-.025em b}\kern-.08em
T\kern-.1667em\lower.7ex\hbox{E}\kern-.125emX}}

\def\volumeyear{2023}

\newcommand{\belSpace}{\ensuremath{\mathcal{B}}\xspace}
\newcommand{\bel}{\ensuremath{b}\xspace}

\newcommand{\stSpace}{\ensuremath{\mathcal{S}}\xspace}

\newcommand{\st}{\ensuremath{s}\xspace}
\newcommand{\stp}{\ensuremath{s'}\xspace}

\newcommand{\actSpace}{\ensuremath{\mathcal{A}}\xspace}
\newcommand{\act}{\ensuremath{a}\xspace}

\newcommand{\obsSpace}{\ensuremath{\mathcal{O}}\xspace}
\newcommand{\obs}{\ensuremath{o}\xspace}

\newcommand{\transF}{\ensuremath{T}\xspace}
\newcommand{\obsF}{\ensuremath{Z}\xspace}

\newcommand{\rewFunc}{\ensuremath{R}\xspace}

\newcommand{\pol}{\ensuremath{\pi}\xspace}
\newcommand{\optPol}{\ensuremath{\pi^*}\xspace}

\newcommand{\belTree}{\ensuremath{\mathcal{T}}\xspace}
\newcommand{\VT}{\ensuremath{\mathcal{H}}\xspace}

\newcommand{\episode}{\ensuremath{e}\xspace}

\newcommand{\cell}{\ensuremath{P}\xspace}

\newcommand{\ccite}[1]{~\citep{#1}}

\newcommand{\fref}[1]{Figure~\ref{#1}}

\newcommand{\tref}[1]{Table~\ref{#1}}

\newcommand{\solver}{{\bf A}daptive {\bf D}iscretization using {\bf V}oronoi {\bf T}rees\xspace}
\newcommand{\solverAbbr}{ADVT\xspace}

\newcommand{\ie}{i.e.\xspace}

\newcommand{\expect}{\ensuremath{\mathbb{E}}}
\newcommand{\del}[1]{\mathrm{d}{#1}}

\DeclareMathOperator{\reals}{\mathbb{R}}

\newcommand{\pomdpTuple}{\ensuremath{\mathcal{P}}\xspace}

\newcommand{\commNew}[1]{{\color{black} #1}}
\newcommand{\reuseTree}{RT}
\newcommand{\bellmanBackup}{BB}

\DeclareMathOperator*{\argmax}{arg\,max}

\begin{document}

\runninghead{Hoerger et al.}

\title{Adaptive Discretization using Voronoi Trees for Continuous POMDPs}

\author{Marcus Hoerger\affilnum{1} and Hanna Kurniawati\affilnum{2} and Dirk Kroese\affilnum{1} and Nan Ye\affilnum{1}}

\affiliation{\affilnum{1}School of Mathematics \& Physics, The University of Queensland, Australia\\
\affilnum{2}School of Computing, Australian National University, Australia}

\corrauth{Marcus Hoerger, School of Mathematics \& Physics, The University of Queensland, Australia}

\email{m.hoerger@uq.edu.au}

\begin{abstract}
Solving continuous Partially Observable Markov Decision Processes (POMDPs) is challenging, particularly for high-dimensional continuous action spaces. To alleviate this difficulty, we propose a new sampling-based online POMDP solver, called \emph{\solver (\solverAbbr)}. It uses Monte Carlo Tree Search in combination with an adaptive discretization of the action space as well as optimistic optimization to efficiently sample high-dimensional continuous action spaces and compute the best action to perform. Specifically, we adaptively discretize the action space for each sampled belief using a hierarchical partition called \emph{Voronoi tree}, which is a Binary Space Partitioning that implicitly maintains the partition of a cell as the Voronoi diagram of two points sampled from the cell. \solverAbbr uses the estimated diameters of the cells to form an upper-confidence bound on the action value function within the cell, guiding the Monte Carlo Tree Search expansion and further discretization of the action space. This enables \solverAbbr to better exploit local information with respect to the action value function, allowing faster identification of the most promising regions in the action space, compared to existing solvers. Voronoi trees keep the cost of partitioning and estimating the diameter of each cell low, even in high-dimensional spaces where many sampled points are required to cover the space well. \solverAbbr additionally handles continuous observation spaces, by adopting an observation progressive widening strategy, along with a weighted particle representation of beliefs. Experimental results indicate that \solverAbbr scales substantially better to high-dimensional continuous action spaces, compared to state-of-the-art methods.
\end{abstract}


\maketitle

\section{Introduction}\label{s:Intro}
Planning in scenarios with non-deterministic action effects 
and partial observability is an essential, yet
challenging problem for autonomous robots. The Partially Observable Markov
Decision Process (POMDP)\ccite{kaelbling1998planning,Sondik:71} is a general principled framework for
such planning problems. POMDPs lift the planning problem from the state
space to the \textit{belief space} --- that is, the set of all probability
distributions over the state space. By doing so, POMDPs enable robots to
systematically account for uncertainty caused by stochastic actions and
incomplete or noisy observations in computing the optimal strategy. Although
computing the optimal strategy exactly is intractable in general\ccite{papadimitriou1987complexity}, the past two decades have seen a surge
of sampling-based POMDP solvers (reviewed in\ccite{Kur22:Partially}) that trade optimality with approximate
optimality for computational tractability, enabling POMDPs to become practical for a variety of realistic robotics problems.

Despite these advances, POMDPs with continuous state, action and observation spaces remain a challenge, particularly for high-dimensional continuous action spaces. 
Recent solvers for continuous-action POMDPs\ccite{fischer2020information,mern2021bayesian,seiler2015online,sunberg2018online} are generally online ---that is, planning and execution are interleaved--- and exploit Monte Carlo Tree Search (MCTS) to find the best action among a finite representative subset of the action space. MCTS interleaves guided belief space sampling, value estimation and action subset refinement to incrementally improve the possibility that the selected subset of actions contains the best action.
They generally use UCB1\ccite{Aue02:Finite} to guide belief space sampling and Monte Carlo backup for value estimation, but differ in the action subset refinement.  


Several approaches use the Progressive Widening strategy\ccite{couetoux2011continuous} to continuously add new randomly sampled actions once current actions have been sufficiently explored. Examples include POMCPOW\ccite{sunberg2018online} and IPFT\ccite{fischer2020information}. More recent algorithms combine Progressive Widening with more informed methods for adding new actions: VOMCPOW\ccite{lim2020voronoi} uses Voronoi Optimistic Optimization\ccite{kim2020monte} and BOMCP\ccite{mern2021bayesian} uses Bayesian optimization. All of these solvers use UCT-style simulations and Monte Carlo backups. An early line of work, GPS-ABT\ccite{seiler2015online}, takes a different approach: It uses Generalized Pattern Search to iteratively select an action subset that is more likely to contain the best action and add it to the set of candidate actions. GPS-ABT uses UCT-style simulations and Bellman backup (following the implementation of ABT\ccite{hoerger2018software,klimenko2014tapir}), though the distinction between Monte Carlo and Bellman backup was not  clarified nor explored.  All of these solvers have been successful in finding good solutions to POMDPs with continuous action spaces, though for a relatively low ($\le 4)$ dimension. 

To compute good strategies for POMDPs with high-dimensional action spaces, we propose a new \commNew{MCTS-based} online POMDP solver, called \solver (\solverAbbr). \commNew{\solverAbbr is designed for problems with continuous action spaces, while the state and observation spaces can either be discrete or continuous.}
\commNew{\solverAbbr contains three key ideas, as briefly described below.

The first key idea of \solverAbbr is a new data structure for action space discretization, called \emph{Voronoi tree}. A Voronoi tree represents a hierarchical partition of the action space for a single sampled belief. It follows the structure of a Binary Space Partitioning (BSP) tree, but each partitioning hyper-plane is only implicitly maintained and computed based on the Voronoi diagram of a pair of sampled actions, which results in a partitioning that is much more adaptive to the spatial locations of the sampled actions, compared to state-of-the-art methods\ccite{bubeck2011x,lim2020voronoi,mansley2011sample,sunberg2018online,valko2013stochastic}. 
We additionally maintain estimates of the diameters of the cells in the Voronoi tree.
These diameters are used in the other two key ideas, namely, action selection and Voronoi tree refinement, 
as described in the next two paragraphs.
The hierarchical structure of the Voronoi tree allows us to estimate the diameters with an efficient 
sampling-based algorithm that scales well to high-dimensional action spaces.
\Cref{sec:voronoitree} provides a more detailed description on the computational and representational advantages of the 
Voronoi tree.}

\commNew{The second key idea of \solverAbbr is the use of a cell-diameter-aware upper-confidence bound on the values of actions to guide action selection during planning. This bound represents an upper-confidence bound on the values of \emph{all} actions within a cell, based on the estimated value of the corresponding sampled action and the diameter of the cell. 
Our bound is a generalization of a bound that was developed for Lipschitz continuum-arm bandit problems 
\ccite{wang2020towards}.
It is motivated by the observation that in many continuous action POMDPs for robotics problems, the distance between two actions can often be used as an indication of how similar their values are. 
Using this observation, \solverAbbr assumes that the action value for a belief is Lipschitz continuous in the action space,
which in turn allows us to derive the upper bound.
The upper bound helps \solverAbbr to exploit local information 
(i.e., the estimated value of the representative action of a cell and the cell diameter) with respect to the action value function and bias its search towards the most promising regions of the action space. }


\commNew{The third key idea is a diameter-aware cell refinement rule.
We use the estimated diameters of the cells to help \solverAbbr decide if a cell needs further refinement:
a larger cell will be split into smaller ones after a small number of simulations, while a smaller will 
require more simulations before it is split.
This helps \solverAbbr to avoid an unnecessarily small partitioning of non-promising regions in the action space.

To further support \solverAbbr in efficiently finding approximately optimal actions, we use a stochastic version of the Bellman backup\ccite{klimenko2014tapir} rather than the typical Monte-Carlo backup to estimate the values of sampled actions. Stochastic Bellman backups help \solverAbbr to backpropagate the value of good actions deep in the search tree, instead of averaging them out. This strategy of estimating the action values is particularly helpful for problems with sparse rewards.}

\commNew{Aside from continuous action spaces, continuous observation spaces pose an additional challenge for MCTS-based online POMDP solvers. Recent approaches such as POMCPOW and VOMCPOW apply Progressive Widening in the observation space in conjunction with an explicit representation of the sampled beliefs via a set of weighted particles. Due to its simplicity, we adopt POMCPOW's strategy to handle continuous observation spaces. This enables \solverAbbr to scale to problems with continuous state, action and observation spaces.}

Experimental results on a variety of benchmark problems with increasing dimension (up to 12-D) of the action space \commNew{and problems with continuous observation spaces} indicate that \solverAbbr substantially outperforms state-of-the-art methods\ccite{lim2020voronoi,sunberg2018online}. Our C++ implementation of ADVT is available at \url{https://github.com/hoergems/ADVT}.

\commNew{This paper extends our previous work\ccite{hoerger22:ADVT} in three ways: First is the extension of \solverAbbr to handle continuous observation spaces. Experimental results on an additional POMDP benchmark problem demonstrate the effectiveness of \solverAbbr in handling purely continuous POMDP problems. Second is a substantially expanded discussion on the technical concepts of \solverAbbr. And third is an extended ablation study in which we investigate the effectiveness of stochastic Bellman backups when applied to two baseline solvers POMCPOW and VOMCPOW.}

\section{Background and Related Work}

A POMDP provides a general mathematical framework for sequential decision making
under uncertainty.
Formally, it is an 8-tuple 
$\langle \stSpace, \actSpace, \obsSpace, \transF, \obsF, \rewFunc, \bel_{0}, \gamma \rangle$.
The robot is initially in a hidden state $s_{0} \in \stSpace$. 
This uncertainty is represented by an initial belief $\bel_0 \in \belSpace$, which is a probability distribution on the state space $\stSpace$, \commNew{where \belSpace is the set of all possible beliefs}.
At each step $t \ge 0$, the robot executes an action $\act_{t} \in \actSpace$
according to some policy $\pol$.
It transitions to a next state $\st_{t+1} \in \stSpace$ \commNew{according to the transition model} $\transF(\st_{t}, \act_{t}, \st_{t+1}) = p(\st_{t+1} | \st_{t}, \act_{t})$. \commNew{For discrete state spaces, $\transF(\st_{t}, \act_{t}, \st_{t+1})$ represents a probability mass function, whereas for continuous state spaces, it represents a probability density function.} 
The robot does not know the state $\st_{t+1}$ exactly, but perceives an observation $\obs_{t} \in\obsSpace$ \commNew{according to the observation model} 
$\obsF(\st_{t+1}, \act_{t}, \obs_{t}) = p(\obs_{t} | \st_{t+1}, \act_{t})$. \commNew{$\obsF(\st_{t+1}, \act_{t}, \obs_{t})$ represents a probability mass function for discrete observation spaces, or a probability density function for continuous observation spaces respectively.}
In addition, it receives an immediate reward $r_{t} = R(\st_{t}, \act_{t}) \in \reals$.
The robot's goal is to find a policy $\pol$ that maximizes the expected total
discounted reward or the policy value
\begin{align}
    V_{\pol}(\bel_{0}) = \expect\left[\sum_{t=0}^{\infty}\gamma^t r_t \bigg| \bel_{0}, \pi\right],
\end{align}
where the discount factor $0 < \gamma < 1$ ensures that $V_{\pol}(\bel)$ is
finite and well-defined.

The robot's decision space is the set $\Pi$ of policies defined as mappings
from beliefs to actions. 
The POMDP solution is then the optimal policy, denoted as \optPol and defined as  
\begin{align}
\optPol = \argmax_{\pol \in \Pi} V_{\pol}(\bel).
\end{align}
In designing solvers, it is often convenient to work with the action value or $Q$-value 
\begin{align}
Q(\bel, \act) 
= R(\bel, \act) 
+ 
\gamma \mathbb{E}_{\obs\in\obsSpace}[V_{\pol^{*}}(\bel_{\act}^{\obs}) | \bel],
\end{align}
where $R(\bel, \act) = \int_{\st\in\stSpace}\bel(\st)R(\st, \act)\del\st$ is the expected reward of executing action $\act$ at belief $\bel$, while  $\bel_{\act}^{\obs} = \tau(\bel, \act, \obs)$ is the updated robot's belief estimate after it performs action ${\act \in \actSpace}$  while at belief \bel, and subsequently perceives observation $\obs \in \obsSpace$. The optimal value function is then 
\begin{align}
    V^*(\bel) = \max_{\act\in\actSpace} Q(\bel, \act).
\end{align} 
A more elaborate explanation is available in\ccite{kaelbling1998planning}.

Belief trees are convenient data structures to find good approximations  to the optimal solutions via sampling-based approaches, which has been shown to significantly improve the scalability of POMDP solving\ccite{Kur22:Partially}. Each node in a belief tree represents a sampled belief. It has outgoing edges labelled by actions, and each action edge is followed by outgoing edges labelled by observations which lead to updated belief nodes. Na\"ively, bottom-up dynamic programming can be applied to a truncated belief tree to obtain a near-optimal policy, but many scalable POMDP solvers use more sophisticated \commNew{sampling} strategies to construct a compact belief tree, from which a close-to-optimal policy can be computed efficiently. \solverAbbr uses such a sampling-based approach and belief tree representation too.

Various efficient sampling-based offline and online POMDP solvers have been developed for increasingly complex discrete and continuous POMDPs in the last two decades. Offline solvers (e.g.,  \citealt{bai2014integrated,kurniawati2011motion,Kurniawati08sarsop:efficient,Pin03:Point,Smi05:Point}) compute an approximately optimal policy for all beliefs first before deploying it for execution. In contrast, online solvers (e.g., \citealt{kurniawati2016online,silver2010monte,somani2013despot}) aim to further scale to larger and more complex problems by interleaving planning and execution, and focusing on computing an optimal action for only the current belief during planning. For scalability purposes, \solverAbbr follows the online solving approach.

Some online solvers have been designed for continuous POMDPs. In addition to the general solvers discussed in \Cref{s:Intro}, some solvers\ccite{agha2011firm,sun2015high,van2011lqg,van2012motion} restrict beliefs to be Gaussian and use Linear-Quadratic-Gaussian (LQG) control\ccite{Lindquist73} to compute the best action. This strategy generally performs well in high-dimensional action spaces. However, they tend to perform poorly in problems with large uncertainties\ccite{hoerger2020linearization}.


Last but not least, hierarchical rectangular partitions have been commonly \commNew{used to discretize action spaces} when solving continuous action bandits and MDPs (the fully observed version of POMDPs), such as HOO\ccite{bubeck2011x} and HOOT\ccite{mansley2011sample}. However, the partitions used in these algorithms are typically predefined, which are less adaptive than Voronoi-based partitions constructed dynamically during the search. On the other hand, Voronoi partitions have been proposed in VOOT\ccite{kim2020monte} and VOMCPOW\ccite{lim2020voronoi}. However, their partitions are based on the Voronoi diagram of all sampled actions, which makes the computation of cell diameters and sampling relatively complex in high-dimensional action spaces. 
\solverAbbr is computationally efficient, just like hierarchical rectangular partitions, and yet adaptive, just like the Voronoi partitions, getting the best of both worlds.

\section{\solverAbbr: Overview}
\begin{algorithm}[tbp]
\caption{\textproc{\solverAbbr}(Initial belief $\bel_0$)}\label{alg:mcts}
\begin{algorithmic}[1]
\State $\bel = \bel_0$ 
\State $\belTree = $\ initializeBeliefTree($\bel$) 
\State $\VT(\bel) = $ Initialize Voronoi tree for belief \bel
\State isTerminal $=$\ False
\While{isTerminal is False}
    \While{planning budget not exceeded}
        \State $(\episode, \bel_c) = $\ \textproc{SampleEpisode}(\belTree, \bel)\Comment{\Cref{alg:sampleEpisode}}        
        \For{$i = \left |\episode \right | - 1$ to 1}
               \State $(\st, \act, \obs, r) = \episode_i$
		\State \textproc{Backup}(\belTree, $\bel_c$, \act, $r$) \Comment{\Cref{alg:backup}}
		\State $\bel_c = $\ Parent node of $\bel_c$ in \belTree
		\State \textproc{RefineVoronoiTree}($\VT(\bel_c)$, \act) \Comment{\Cref{alg:updateAction}}		
        \EndFor
    \EndWhile     
    \State $\act^* = \argmax_{\act\in\actSpace(\bel)} \widehat{Q}(\bel, \act)$
    \State $(\obs$,\ isTerminal$) =$\ Execute $\act^*$    
    \State $\bel' = \tau(\bel, \act^*, \obs)$ 
    \State $\bel = \bel'$    
\EndWhile
\end{algorithmic}
\end{algorithm}

In this paper, we consider a POMDP $\pomdpTuple = \langle \stSpace, \actSpace, \obsSpace, \transF, \obsF, \rewFunc, \bel_{0}, \gamma \rangle$, where the action space \actSpace is continuous and embedded in a bounded metric space with distance function $d$. Typically, we define the metric space to be a $D$-dimensional bounded Euclidean space, though \solverAbbr can also be used with other types of bounded metric spaces. We further consider the state space \stSpace and observation space \obsSpace to be either discrete or continuous. 

\solverAbbr assumes that the $Q$-value \commNew{function} is Lipschitz continuous in the action space; that is, for any belief $\bel \in \belSpace$, there exists a Lipschitz constant $L_b$ such that for any actions $\act, \act' \in \actSpace$, we have $\left |Q(\bel, \act) - Q(\bel, \act') \right | \leq L_b\ d(\act, \act')$. 
Since generally we do not know a tight Lipschitz constant, in the implementation, \solverAbbr uses the same Lipschitz constant $L$ for all beliefs in \belSpace, as discussed in \Cref{ssec:actSelection}.

\solverAbbr is an anytime online solver for POMDPs. 
It interleaves belief space sampling and action space sampling to find the best action to perform from the current belief $\bel \in \belSpace$. 
The sampled beliefs are maintained in a belief tree, denoted as \belTree, while the sampled actions $\actSpace(b)$ 
for a belief $b$ are maintained in a Voronoi tree, denoted as $\VT(b)$, \commNew{which is adaptively refined}. 
The Voronoi trees form part of the belief tree in \solverAbbr:
they determine the sampled action branches for the belief nodes.

\Cref{alg:mcts} presents the overall algorithm of \solverAbbr, with details in the sections below. \commNew{At each planning step, \solverAbbr follows the MCTS approach of constructing a belief tree by sampling a set of \textit{episodes} (line 7 in \Cref{alg:mcts}), starting from the current belief. Details on the belief-tree construction are provided in \Cref{ssec:beliefTree}. After sampling an episode, the estimated action values $\widehat{Q}(\bel, \act)$ along the sequence of actions selected by the episode are updated using a backup operation (line 10 in \Cref{alg:mcts}). In addition, \solverAbbr refines the Voronoi tree $\VT(\bel)$ as needed for each belief visited by the episode (line 12 in \Cref{alg:mcts}), as discussed in \Cref{sec:voronoitree}. Once a planning budget is exceeded, \solverAbbr selects an action $\act^*$ from the current belief acoording to 
\begin{equation}
\act^* = \argmax_{\act\in\actSpace(\bel)}\widehat{Q}(\bel, \act),
\end{equation}
executes $\act^*$ in the environment to obtain an observation $\obs\in\obsSpace$, updates the current belief to $\bel' = \tau(\bel, \act^*, \obs)$ (line 15-17 in \Cref{alg:mcts}), and proceeds planning from the updated belief. This process repeats until the robot enters a terminal state or a maximum number of planning steps has been exceeded. 
}

\section{\solverAbbr: Construction of the Belief Tree}
\label{ssec:beliefTree}

\begin{algorithm}[htbp]
\caption{\textproc{SampleEpisode}(Belief tree \belTree, Belief node $\bel_c$)}\label{alg:sampleEpisode}
\begin{algorithmic}[1]
\State Notations: $\VT(\bel) =$\ Voronoi tree associated with belief \bel; $A(b) =$\ Set of candidate actions associated to the leaf-nodes of $\VT(\bel)$
\item[]
\State $\episode = $\ Empty sequence of state-action-observation-reward quadruples; $\bel = \bel_c$; $\st = $\ A random state sampled from \bel; newBelief $=$ False
\While{newBelief is False and \st not terminal}
    \State $\act = \argmax_{\act_k\in\actSpace(\bel)} U(\bel, \act_k)$
\Comment{\cref{eq:action_selection}}
  \State $(\bel', \stp, \obs, r) =$\ \textproc{Step}(\bel, \st, \act) \Comment{\Cref{alg:observationSampling}}
  \State Append $(\st, \act, \obs, r)$ to $\episode$
  \State $N(\bel, \act) = N(\bel, \act) + 1; N(\bel) = N(\bel) + 1$ 
  \If {$\actSpace(\bel') = \emptyset$}  
  	\State $\act =\ $Sample uniformly from \actSpace	
  	\State $\VT(\bel') = $ Initialize Voronoi tree for belief $\bel'$  	
  	\State Associate $(\act, \actSpace)$ with the root node of $\VT(\bel')$   
  	\State $N(\bel') = 0$; $N(\bel', \act) = 0$
  	\State newBelief = True
  \EndIf
  \State $\st = \stp$ 
  \State $\bel = \bel'$
\EndWhile
\State $r = 0$
\If{newBelief is True}
	\State $h = $\ calculateRolloutHeuristic(\st, \bel)
  	\State Initialize $\widehat{V}^*(\bel)$ with $h$
\EndIf
\State insert $(s, -, -, r)$ to $\episode$
\State \Return $(\episode, \bel)$
\end{algorithmic}
\end{algorithm}

The belief tree \belTree is a tree whose nodes represent beliefs and the edges are associated with action--observation pairs $(\act, \obs)$, where $\act \in \actSpace$ and $\obs \in \obsSpace$. A node $\bel'$ is a child of node $\bel$ via edge $(\act, \obs$) if and only if $\bel' = \tau(\bel, \act, \obs)$.


To construct the belief tree \belTree, \solverAbbr interleaves the iterative
select-expand-simulate-backup operations used in many MCTS algorithms with
adaptive discretization. \commNew{We assume that we have access to a \textit{generative model} $G: \stSpace \times \actSpace \rightarrow \stSpace \times \obsSpace \times \mathbb{R}$ that simulates the dynamics, observation and reward models. In particular, for a given state $\st\in\stSpace$ and action $\act\in\actSpace$, we have that $(\st', \obs, r) = G(\st, \act)$, where $(\st', \obs)$ is distributed according to $p(\st', \obs | \st, \act)=T(\st, \act, \st')Z(\st', \act, \obs)$, and $r=R(\st, \act)$.}
At each iteration, \commNew{we} first \emph{select} a path starting from the root by sampling
an episode 
$\st_{0}, \act_{0}, \obs_{0}, r_{0}, \st_{1}, \act_{1}, \obs_{1}, r_{1},
\ldots$ as follows:
\commNew{We first} set the current node $\bel$ as the root node, and sample $\st_{0}$ from $\bel$.
At each step $i \ge 0$, we choose an action $\act_{i} \in \actSpace(\bel)$ for
\bel using an action selection strategy (discussed in \Cref{ssec:actSelection})\commNew{, execute $\act_{i}$ from state $\st_i$ via the generative model $G$ to obtain a next state $\st'$, an observation \obs and the immediate reward $r_i$. For problems with discrete observation spaces, we set the episode's next state $\st_{i+1}$ and observation $\obs_i$ to $\st_{i+1} = \st'$ and $\obs_{i} =\obs$ respectively. For problems with continuous observation spaces, we select $\st_{i+1}$ and $\obs_i$ according to an observation sampling strategy, as discussed in \Cref{ssec:observation_sampling_strategy}}. Finally, we update \bel to $\bel$'s child node via $(\act_{i}, \obs_{i})$. The process terminates when encountering a terminal state or when the child node
does not exist; in the latter case, the tree is \emph{expanded} by adding a new node,
and a rollout policy is \emph{simulated} to provide an estimated value for the new
node.
In either case, \emph{backup} operations are performed to update the estimated
values for all \commNew{actions selected by the episode}.
\commNew{In addition, new actions are added to $\actSpace(\bel)$ by refining the associated Voronoi tree $\VT(\bel)$ as needed for each encountered belief.} 
\Cref{alg:sampleEpisode} presents the pseudo-code for constructing \belTree, while the backup operation and refinement of \commNew{$\VT(\bel)$} are discussed in \Cref{ssec:backup} and \Cref{sec:voronoitree}, respectively.

\subsection{Action Selection Strategy}
\label{ssec:actSelection}

In contrast to many existing online solvers, which use UCB1 to select the action to expand a node \bel of \belTree, \solverAbbr treats the action selection problem as a continuum-arm bandit problem. Specifically, it selects an action from the set of candidate actions $\actSpace(\bel)$ according to\ccite{wang2020towards}
\begin{align}
	\act^{*} &= \argmax_{\act \in\actSpace(\bel)} U(\bel, \act), \qquad\text{with} 
	    \label{eq:action_selection} \\
    U(\bel, \act) &= \widehat{Q}(\bel, \act) + C\sqrt{\frac{\log N(\bel)}{N(\bel, \act)}} + L\ \mathrm{diam}(\cell),
        \label{eq:u_value}
\end{align}
where $N(\bel)$ is the number of times node \bel has been visited so far, $N(\bel, \act)$ is the number of times $\act$ has been selected at \bel, $\cell \subseteq \actSpace$ is the unique leaf cell containing $\act$ in
$\VT(\bel)$ (see \Cref{sec:voronoitree} for details on the Voronoi tree),
and $\mathrm{diam}(\cell) = \sup_{\act, \act' \in \cell} d(\act, \act')$ is the
diameter of $\cell$ with respect to the distance metric $d$.
The constant $C$ is an exploration constant, where larger values of $C$ encourage exploration. 
In case $N(\bel, \act) = 0$, we set $U(\bel, \act) = \infty$. 
With the Lipschitz continuity assumption, the value $U(\bel, \act)$ can be seen as an upper-confidence bound for
the maximum possible Q-value $\max_{\act' \in \cell} Q(\bel, \act')$ within $P$, as follows:
$\widehat{Q}(\bel, \act) + C\sqrt{\frac{\log N(\bel)}{N(\bel, \act)}}$ is 
the standard UCB1 bound for the Q-value $Q(b, \act)$, and whenever this 
upper bounds $Q(\bel, \act)$, we have 
$U(\bel, \act) \ge Q(\bel, \act')$ for any $\act' \in \cell$, because
$U(\bel, \act) 
\ge Q(\bel, \act) + L\ \mathrm{diam}(\cell) 
\ge Q(\bel, \act')$, where the last inequality holds due to the Lipschitz assumption.
Since $L$ is unknown, we try different values of $L$ in our experiments and choose the best.

\subsection{Backup}
\label{ssec:backup}


\begin{algorithm}[htbp]
\caption{\textproc{Backup}(Belief tree \belTree, Belief node $\bel'$, Action \act, Reward $r$)}\label{alg:backup}
\begin{algorithmic}[1]
        \State $\bel = $\ Parent node of $\bel'$ in \belTree
	\State {$\widehat{Q}(\bel, \act) \gets \widehat{Q}(\bel, \act) + \frac{r + \gamma \widehat{V}^*(\bel') - \widehat{Q}(\bel, \act)}{N(\bel, \act)}$}
	\State {$\widehat{V}^*(\bel) = \max_{\act\in\actSpace(\bel)} \widehat{Q}(\bel, \act)$}
\end{algorithmic}
\end{algorithm}
After sampling an episode \episode, \solverAbbr updates the estimates $\widehat{Q}(\bel, \act)$ as well as the statistics $N(\bel)$ and $N(\bel, \act)$ along the sequence of beliefs visited by the episode. To update $\widehat{Q}(\bel, \act)$, we use a stochastic version of the Bellman backup (\Cref{alg:backup}): Suppose $r$ is the immediate reward sampled by the episode after selecting \act from \bel. We then update $\widehat{Q}(\bel, \act)$ according to 
\begin{equation}\label{eq:\bellmanBackup}
\widehat{Q}(\bel, \act) \gets \widehat{Q}(\bel, \act) + \frac{r + \gamma \widehat{V}^*(\bel') - \widehat{Q}_i(\bel, \act)}{N(\bel, \act)},
\end{equation}
where $\bel'$ is the child of \bel in the belief tree \belTree via edge $(\act, \obs)$; \ie, the belief we arrived at after performing action $\act \in \actSpace$ and perceiving observation ${\obs \in \obsSpace}$ from \bel, and $\widehat{Q}_i(\bel, \act)$ is the previous estimate of $Q(\bel, \act)$. This rule is in contrast to POMCP, POMCPOW and VOMCPOW, where the $Q$-value estimates are updated via Monte Carlo backup, \ie, \commNew{
\begin{equation}\label{eq:mcBackup}
\widehat{Q}(\bel, \act) \gets \widehat{Q}(\bel, \act) + \frac{r + \gamma \widehat{V}_{\episode}(\bel') - \widehat{Q}_i(\bel, \act)}{N(\bel, \act)},
\end{equation}
where $\widehat{V}_{\episode}(\bel')$ is the the total discounted reward of episode \episode, starting from belief $\bel'$.} 
 
Our update rule \commNew{in \cref{eq:\bellmanBackup}} is akin to the rule used in $Q$-Learning\ccite{watkins1992q} and was implemented in the ABT software\ccite{hoerger2018software,klimenko2014tapir}, though never explicitly compared with Monte Carlo backup. 

The update rule in \cref{eq:\bellmanBackup} helps \solverAbbr to focus its search on promising parts of the belief tree, particularly for problems where good rewards are sparse. \commNew{For sparse-reward problems, the values of good actions deep in the search tree tend to get averaged out near the root when using Monte-Carlo backups, thus their influence on the action values at the current belief diminishes. In contrast, since stochastic Bellman backups always backpropagates the current largest action value for a visited belief, large action values deep in the search tree have a larger influence on the action values near the root. As we will demonstrate in the experiments, using stochastic Bellman backups instead of Monte-Carlo backups can lead to a significant performance benefit for many POMDP problems.}

\subsection{Observation Sampling Strategy}\label{ssec:observation_sampling_strategy}
\commNew{
\begin{algorithm}[!htbp]
\caption{\textproc{Step}(Belief node \bel, State \st, Action \act)}\label{alg:observationSampling}
\begin{algorithmic}[1]
        \State $(\stp, \obs, r) = G(\st, \act)$ \Comment{Generative model}
        \State $\bel' = null$        
        \If{$\obsSpace$ is discrete}
            \State $\bel' =$\ Child node of \bel via edge $(\act, \obs)$. (If no such child exists, create one)
        \Else
            \If{$\left |\obsSpace(\bel, \act) \right |> k_\obs N(\bel, \act)^{\alpha_\obs}$}
                \State $\obs =$\ Sample \obs uniformly at random from $\obsSpace(\bel, \act)$
            \EndIf
            \State $\bel' =$\ Child node of \bel via edge $(\act, \obs)$. (If no such child exists, create one)
            \State $w = Z(\stp, \act, \obs)$
            \State $\bel' \cup \{(\stp, w) \}$
            \State $\stp\sim\bel'$ 
            \State $r = R(\st, \act, \stp)$                        
        \EndIf
        \State \Return $(\bel', \stp, \obs, r)$        
\end{algorithmic}
\end{algorithm}

During the episode-sampling process, when ADVT selects an action \act at a belief \bel according to \cref{eq:action_selection}, we must sample an observation to advance the search to the next belief. ADVT uses different observation sampling strategies, depending on whether the observation space \obsSpace is discrete or continuous. \Cref{alg:observationSampling} summarizes \solverAbbr's observation sampling strategy for both discrete and continuous observation spaces. 

For discrete observation spaces, ADVT follows the common strategy of sampling an observation from the generative model, given the current state of the episode and the selected action, and representing each sampled observation via an observation edge in the search tree. Since the number of possible observations is finite in the discrete setting, this strategy typically works well for moderately sized observation spaces.

For continuous observation spaces, however, the above strategy is unsuitable. In this setting, each sampled observation is generally unique, which leads to a possibly infinite number of observations that need to be represented in the search tree. As a consequence, we cannot expand the search beyond the first step, resulting in policies that are too myopic. Thus, to handle continuous observation spaces, we adopt a strategy similar to the one used by POMCPOW\ccite{sunberg2018online} and VOMCPOW\ccite{lim2020voronoi}. This strategy consists of two components. The first component is to apply Progressive Widening\ccite{couetoux2011continuous} to limit the number of sampled observation edges per action edge as a function of $N(\bel, \act)$, \ie, the number of times we selected \act from \bel. In particular, let $\obsSpace(\bel, \act)$ be the set of observation children of action \act at belief \bel. We sample a new observation as a child $\act$ whenever $|\obsSpace(\bel, \act)|$ satisfies $|\obsSpace(\bel, \act)| \leq k_\obs N(\bel, \act)^{\alpha_\obs}$, where $k_\obs \geq 0$ and $\alpha_\obs \geq 0$ are user defined parameters that control the rate at which new observations are added to the tree. If this condition is not satisfied, we uniformly sample an observation from $\obsSpace(\bel, \act)$. This enables \solverAbbr to visit sampled observation edges multiple times and expand the search tree beyond one step. The second component is to use an explicit representation of each sampled belief in the search tree via a set of weighted particles $\left \{ (\st_i, w_i)_{i=1}^n \right \}$, which allows us to obtain state samples that are correctly distributed according to the beliefs. When sampling an episode, suppose ADVT selects action \act from belief \bel, samples a next state \stp from the generative model and selects an observation \obs using the strategy above. The weight corresponding to \stp is then computed according to $w = Z(\stp, \act, \obs)$ and $(\stp, w)$ is added to particle set representing belief $\bel'$ whose parent is \bel via the edge $(\act, \obs)$. To obtain a next state \stp that is distributed according to $\bel'$, we resample $\stp$ from $\bel'$ with a probability proportional to the particle weights and continue from $\bel'$.

Note that in contrast to online POMDP solvers for discrete observation spaces that only require a black-box model to sample observations, we additionally require access to the observation function $Z(\st, \act, \obs)$. However, this is a common assumption for solvers that are designed for continuous observation spaces\ccite{sunberg2018online, hoerger2021continuous}. 

Additionally, note that it is possible to use the observation sampling strategy for continuous observation spaces for discrete observation spaces. However, for discrete observation spaces, this introduces unneccessary computational overhead. As discussed above, the purpose of explicit belief representations is to obain state samples that are correctly distributed according to the beliefs. For discrete observation spaces, this is achieved by sampling a next state $\stp$ according to the transition model $T(\st, \act, \stp)$ and an observation $\obs$ according to the observation model $Z(\stp, \act, \obs)$, and then use \stp as a sample from the belief $\bel' = \tau(\bel, \act, \obs)$. Since the sampled observations are always distributed according to $Z(\stp, \act, \obs)$, \stp is a sample of $\bel'$. This is in contrast to continuous observation spaces, where \solverAbbr samples observations from a distribution that is potentially different to $Z(\stp, \act, \obs)$ (line 7 in \Cref{alg:observationSampling}). Therefore, \solverAbbr handles discrete and continuous observation spaces separately and avoids the weighting an resampling step in the discrete case, thus saving computation time.  


}

\section{\solverAbbr: Construction and Refinement of Voronoi Trees}
\label{sec:voronoitree}
\begin{algorithm}[!htbp]
\caption{\textproc{RefineVoronoiTree}(Voronoi tree $\VT(\bel)$, Action \act)}\label{alg:updateAction}
\begin{algorithmic}[1]
        \State $(a, \cell) = $\ leaf node of $\VT(\bel)$ with its action component being $\act$
        \If{$C_{r}N(\bel, \act) \geq 1/\mathrm{diam}(\cell)^2$}
            \State $\act'$ = sample from $\cell$
			\State $(\cell_{1}, \cell_{2}) = $\ Child cells of $\cell$ induced by \act and $\act'$
			\State Compute diameters of $\cell_{1}$ and $\cell_{2}$
			\State Add $(\act, \cell_{1})$ and $(\act', \cell')$ as $(\act, \cell)$'s children
        \EndIf	
\end{algorithmic}
\end{algorithm}


For each belief node $\bel$ in the belief tree, its Voronoi tree $\VT(\bel)$ is
a BSP tree for $\actSpace$. Each node in $\VT(\bel)$ consists of a pair $(\act,
\cell)$ with $P \subseteq \actSpace$ and $a\in P$ the representative action of $P$,
and each non-leaf node is partitioned into two child nodes.
The partition of each cell in a Voronoi tree is a Voronoi diagram for two
actions sampled from the cell.

\begin{figure*}
\centering
\small
\includegraphics[width=0.8\textwidth]{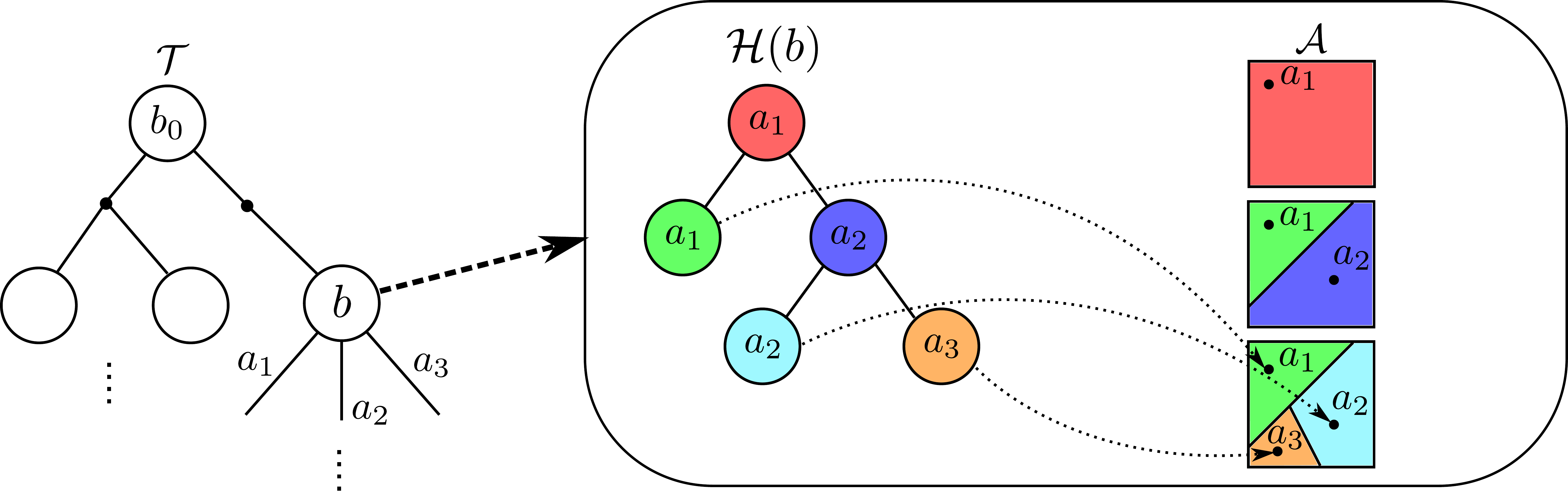}
\caption{Illustration of the relation between a belief tree \belTree (left), the Voronoi tree $\VT(\bel)$ associated to belief \bel (middle) and the partition of the action space induced by the Voronoi tree (right).}
\label{f:Partitioning}
\end{figure*} 


To construct $\VT(\bel)$, \solverAbbr first samples an action $\act_0$ uniformly at random from the action space \actSpace, and sets the pair $(\act_0, \actSpace)$ as the root of $\VT(\bel)$. When \solverAbbr decides to expand a \commNew{leaf node} $(\act, \cell)$, it first samples an action $\act'$ uniformly at random from $\cell \subseteq \actSpace$. \solverAbbr then implicitly constructs the Voronoi diagram between $\act$ and $\act'$ within the cell $\cell$, splitting it into two regions: One is $\cell_1$, representing the set of all actions $\act'' \in \cell$ for which the distance $d(\act'', \act) \leq d(\act'', \act')$, and the other is $\cell_2 = \cell \backslash \cell_1$. The nodes $(\act, \cell_1)$ and $(\act', \cell_2)$ are then inserted as children of $(\act, \cell)$ in $\VT(\bel)$. 
The leaf nodes of $\VT(\bel)$ form the partition of the action space \actSpace used by belief \bel, while the finite action subset $\actSpace(\bel) \subset \actSpace$ used to find the best action from \bel is the set of actions associated with the leaves of $\VT(\bel)$.  \fref{f:Partitioning} illustrates the relationship between a belief, the Voronoi tree $\VT(\bel)$ and the partition of \actSpace.

\commNew{Voronoi trees have a number of representational and computational advantages compared to existing partitioning methods, such as hierarchical rectangular partitions\ccite{bubeck2011x, mansley2011sample} or Voronoi diagrams\ccite{kim2020monte, lim2020voronoi}. In contrast to hierarchical rectangular partitions, Voronoi trees are much more adaptive to the spatial locations of sampled actions, since the geometries of the cells are induced by the sampled actions. Furthermore, in contrast to Voronoi diagrams, the hierarchical structure of Voronoi trees allows us to derive efficient algorithms for estimating the diameters of the cells (discussed in \Cref{ssec:estDiam}) and sampling new actions from a cell (detailed in \Cref{ssec:samplingCells}). For Voronoi diagrams the diameter computation can be prohibitively expensive in the context of online POMDP planning, since each sampled action results in a re-partitioning of the action space; thus, the diameters of the cells have to be re-computed from scratch. Voronoi tress combinine a hierarchical representation of the partition with an implicit construction of the cells via sampled actions, thereby achieving both adaptivity and computational efficiency.}

\commNew{The next section describes how \solverAbbr decides which node of $\VT(\bel)$ to expand.}



\subsection{Refining the Partition}
\label{ssec:refineCell}

\solverAbbr decides how to refine the partitioning $\VT(\bel)$ in two steps. First, it selects a leaf node of $\VT(\bel)$ to be refined next. This step relies on the action selection strategy used for expanding the belief tree \belTree (\Cref{ssec:actSelection}). The selected leaf node $(\act, \cell)$ of $\VT(\bel)$ is the unique leaf node with $\act$ chosen according to \cref{eq:action_selection}

In the second step \solverAbbr decides if the cell \cell\ should indeed be refined. \commNew{This decision is} based on the quality of the estimate $\widehat{Q}(\bel, \act)$, \commNew{as reflected by the number of samples used to estimate $\widehat{Q}(\bel, \act)$}, and the variation of the $Q$-values for the actions contained in \cell, \commNew{as reflected by the diameter of \cell}. Specifically, \solverAbbr refines the cell \cell\ only when the following criteria is satisfied:
\begin{equation}\label{eq:refinement_condition}
    C_{r} N(\bel, \act) \geq \frac{1}{\mathrm{diam}(\cell)^2},
\end{equation}
where $C_{r}$ is an exploration constant. $N(\bel, \act)$, \ie, the number of times that $\act$ has been selected at
$\bel$, provides a rough estimate on the quality of the $\widehat{Q}(\bel, \act)$ estimate, \commNew{whereas $\mathrm{diam}(\cell)$ serves as a measure of variation of the $Q$-value function within \cell due to the Lipschitz assumption}. This criterion, which is inspired by\ccite{touati2020zooming}, limits the growth of the finite set of candidate actions $\actSpace(\bel)$ and ensures that a cell is only refined when its corresponding action has been played sufficiently often. Larger $C_r$ cause cells to be refined earlier, thereby encouraging exploration.

Our refinement strategy is highly adaptive, in the sense that we use local information (i.e., the diameters of the cells, induced by the representative actions \commNew{and the quality of the $Q$-value estimates of the representative actions}), to influence the choice of the cell to be partitioned and when the chosen cell is partitioned, and the geometries of our cells are dependent on the sampled actions. 
This strategy is in contrast to other hierarchical decompositions, such as those used in HOO and HOOT, where the cell that corresponds to an action is refined immediately after the action is selected for the first time, which generally means the $Q$-value of an action is estimated based only on a single play of the action, which is grossly insufficient for our problem.
In addition, our strategy is more adaptive than VOMCPOW. \commNew{For VOMCPOW, the decision on when to refine the partition is solely based on the number of times a belief has been visited and neither takes the quality of the $Q$-value estimates, nor the diameters of the Voronoi cells into account. Furthermore, VOMCPOW's refinement strategy is more global in a sense that each each sampled action results in a different Voronoi diagram of the action space. On the other hand, \solverAbbr's strategy is much more local, since each refinement only affects a single cell.}

\subsection{Estimating the Voronoi Cell Diameters}
\label{ssec:estDiam}

\solverAbbr uses the diameters of the cells in the action selection strategy and the cell refinement rule, but efficiently computing the  diameters of the cells is computationally challenging in high-dimensional spaces.
We give an efficient approximation algorithm for computing the Voronoi cell diameters below.


Since the cells in $\VT(\bel)$ are only implicitly defined, we use a sampling-based approach to approximate a cell's diameter. Suppose we want to estimate the diameter of the cell \cell\ corresponding to the node $(\act, \cell)$ of the Voronoi tree $\VT(\bel)$. Then, we first sample a set of $k$ boundary points ${\actSpace}_{\cell}(\bel)$ of \cell, \commNew{where $k$ is a user defined parameter. In our experiments, we typically set $k$ to be between $20$ and $50$.} To sample a boundary point $\act_\cell \in {\actSpace}_{\cell}(\bel) $, we first sample a point $\alpha$ that lies on the sphere centered at $\act$ with diameter $\mathrm{diam}(\actSpace)$ -- which can be easily computed for our benchmark problems -- uniformly at random. 
The point $\alpha$ lies either on the boundary or outside of \cell. We then use the Bisection method\ccite{Burden2016numerical} with \act and $\alpha$ as the initial end-points, until the  two end-points are less than a small threshold $\epsilon$ away from each other, but one still lies inside \cell\ and the other outside \cell. The point that lies inside \cell\ is then a  boundary point $\act_\cell$. The diameter of a bounding sphere that encloses all the sampled boundary points in ${\actSpace}_{\cell}(\bel)$\ccite{Welzl91enclosing} is then an approximation of the diameter of \cell.

\commNew{The above strategy to estimate the diameter of a cell requires us to determine whether a sampled action \act lies inside or outside a cell \cell. Fortunately, a Voronoi tree allows us to determine if $\act\in\cell$ easily. Observe that for each cell \cell, we have that $\cell \subseteq \mathrm{PARENT}(\cell)$, where $\mathrm{PARENT}(\cell)$ is the cell associated to the parent node of \cell in the Voronoi tree. As a result, $\act\in\cell$ implies that $\act\in\mathrm{PARENT}(\cell)$. Therefore, to check whether $\act\in\cell$, it is sufficent to check if \act is contained in all cells associated to the path in the Voronoi tree from the root to \cell. Suppose $\zeta = \left ( (\act_0, \cell_0), (\act_1, \cell_1), ..., (\act_L, \cell_L)\right )$ is the sequence of nodes corresponding to a path in $\VT(\bel)$ and we want to check whether $\act\in\cell_L$. The point $\act$ is inside $\cell_L$ if, for each $(\act_i, \cell_i)\in\zeta$, we have that $d(\act, \act_i) \leq d(\act, \act'_i)$, where $\act'_i$ is the inducing point of the cell corresponding to the sibling node of $(\act_i, \cell_i)$, and $d(\cdot, \cdot)$ is the distance function on the action space. If, this condition is not satisfied, $\act$ is outside $\cell_L$, \ie, $\act\notin \cell_L$.}

\commNew{To further increase the computational efficiency of our diameter estimator, we re-use the sampled boundary points ${\actSpace}_{\cell}(\bel)$ when \solverAbbr decides to split \cell into two child cells $\cell'$ and $\cell''$. Suppose the diameter of \cell was estimated using $k$ boundary points. Since every point in ${\actSpace}_{\cell}(\bel)$ lies either on the boundary of $\cell'$ or $\cell''$, we divide ${\actSpace}_{\cell}(\bel)$ into ${\actSpace}_{\cell'}(\bel)$ and ${\actSpace}_{\cell''}(\bel)$, such that $\actSpace_{\cell'}(\bel) = \left \{\act \in \actSpace_{\cell}(\bel)\ |\ d(\act, \act') \leq d(\act, \act'') \right \}$ and $\actSpace_{\cell''}(\bel) = \actSpace_{\cell}(\bel)\backslash \actSpace_{\cell'}(\bel)$, where $\act'$ and $\act''$ are the inducing points of $\cell'$ and $\cell''$ respectively. This will leave us with $\left |\actSpace_{\cell'}(\bel) \right | \leq k$ and $\left |\actSpace_{\cell''}(\bel) \right |\leq k$ boundary points for cell $\cell'$ and $\cell''$ respectively. For both $\actSpace_{\cell'}(\bel)$ and $\actSpace_{\cell''}(\bel)$ we then sample $k - \left |\actSpace_{\cell'}(\bel) \right |$ and $k - \left |\actSpace_{\cell''}(\bel) \right |$ additional boundary points using the method above such that $\left |\actSpace_{\cell'}(\bel) \right | = k$ and $\left |\actSpace_{\cell''}(\bel) \right | = k$. Using this method, we only need to sample $k$ new boundary points instead of $2k$ when \solverAbbr decides to split the cell \cell, leading to increased computational efficiency.} 


\subsection{Sampling from the Voronoi Cells}
\label{ssec:samplingCells}

To sample an action that is approximately uniformly distributed in a cell \cell, we use a simple Hit \& Run approach\ccite{smith1984efficient} that performs a random walk within \cell. 
Suppose \cell\ is the cell corresponding to the node $(\act, \cell)$ of the Voronoi tree $\VT(\bel)$. We first sample an action $\act_\cell$ on the boundary of \cell\ using the method described in \Cref{ssec:estDiam}. Subsequently, we take a random step from \act in the direction towards $\act_\cell$, resulting in a new action $\act'\in\cell$. We then use $\act'$ as the starting point, and iteratively perform this process for $m$ steps, which gives us a point that is approximately uniformly distributed in \cell.

\section{Experiments and Results}

We evaluated \solverAbbr on 5 robotics tasks, formulated as continuous-action POMDPs. \commNew{The following section provides details regarding the tasks, while \Cref{t:pomdp_properties} summarizes the state, action and observation spaces for each problem scenario.}

\subsection{Problem Scenarios}
\begin{figure*}[htb]
\centering
\begin{tabular}{c@{\hskip5pt}c@{\hskip5pt}c@{\hskip5pt}c@{\hskip5pt}c}
\includegraphics[height=0.165\textwidth]{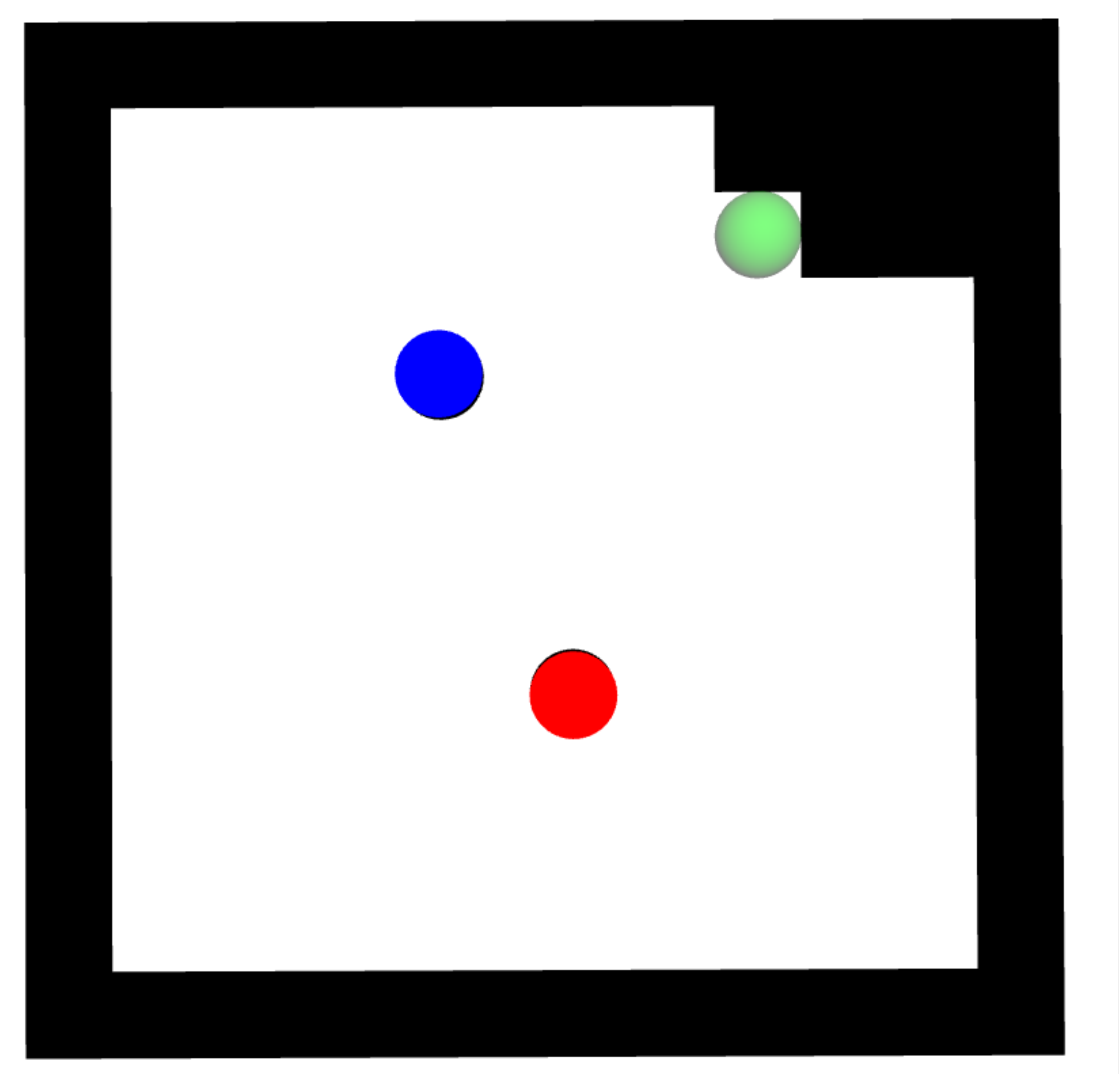} &
\includegraphics[height=0.165\textwidth]{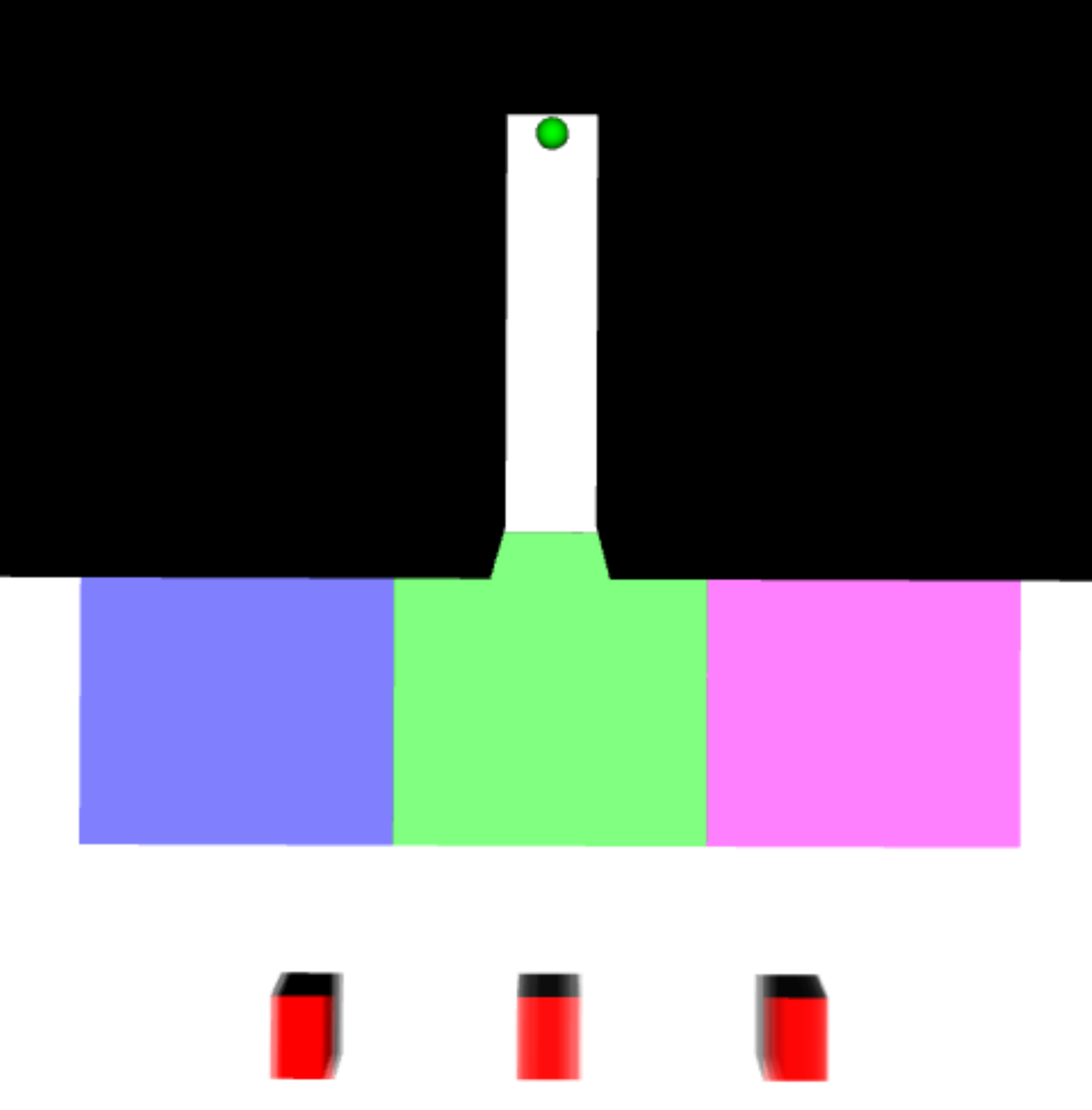} &
\includegraphics[height=0.165\textwidth]{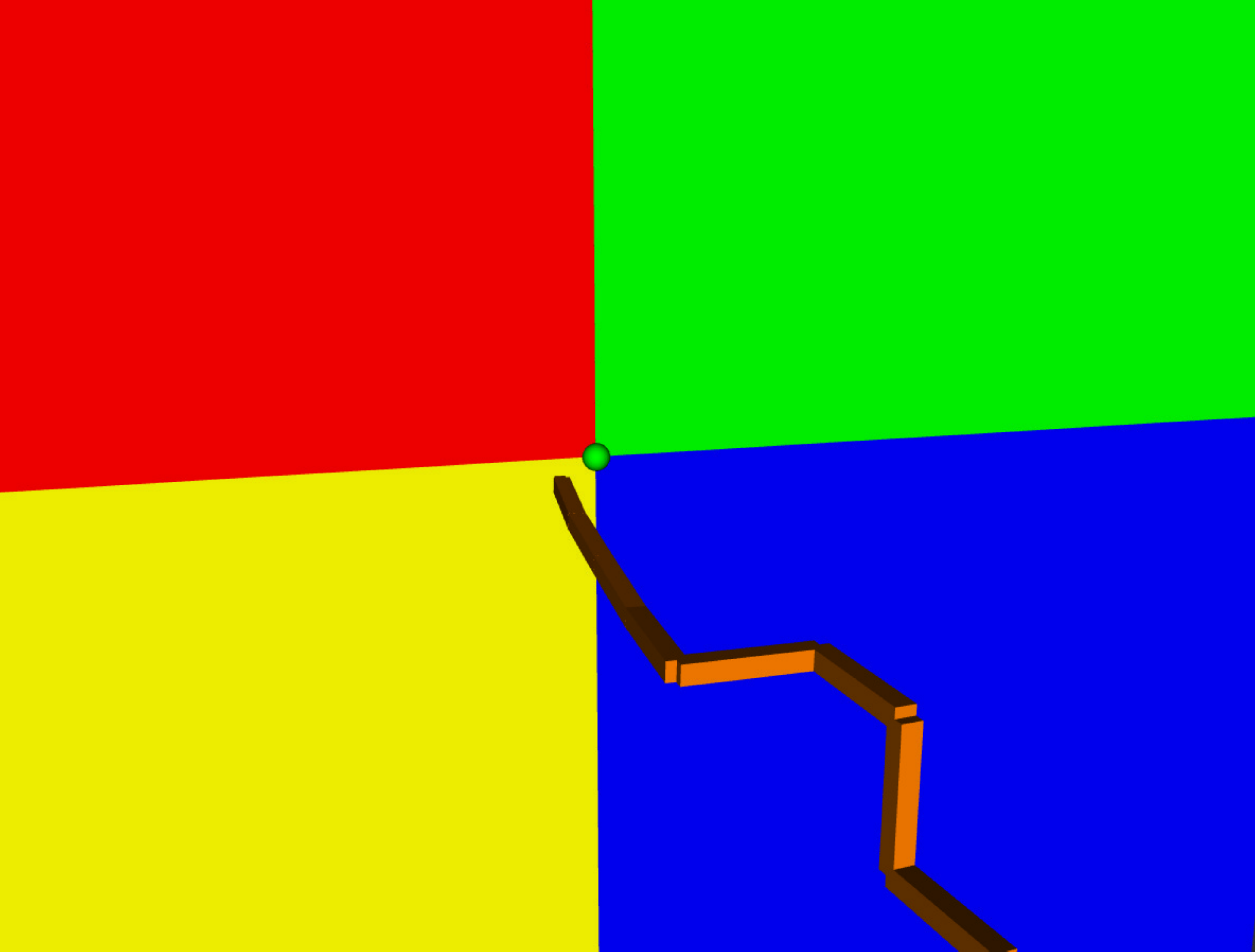} &
\includegraphics[height=0.165\textwidth]{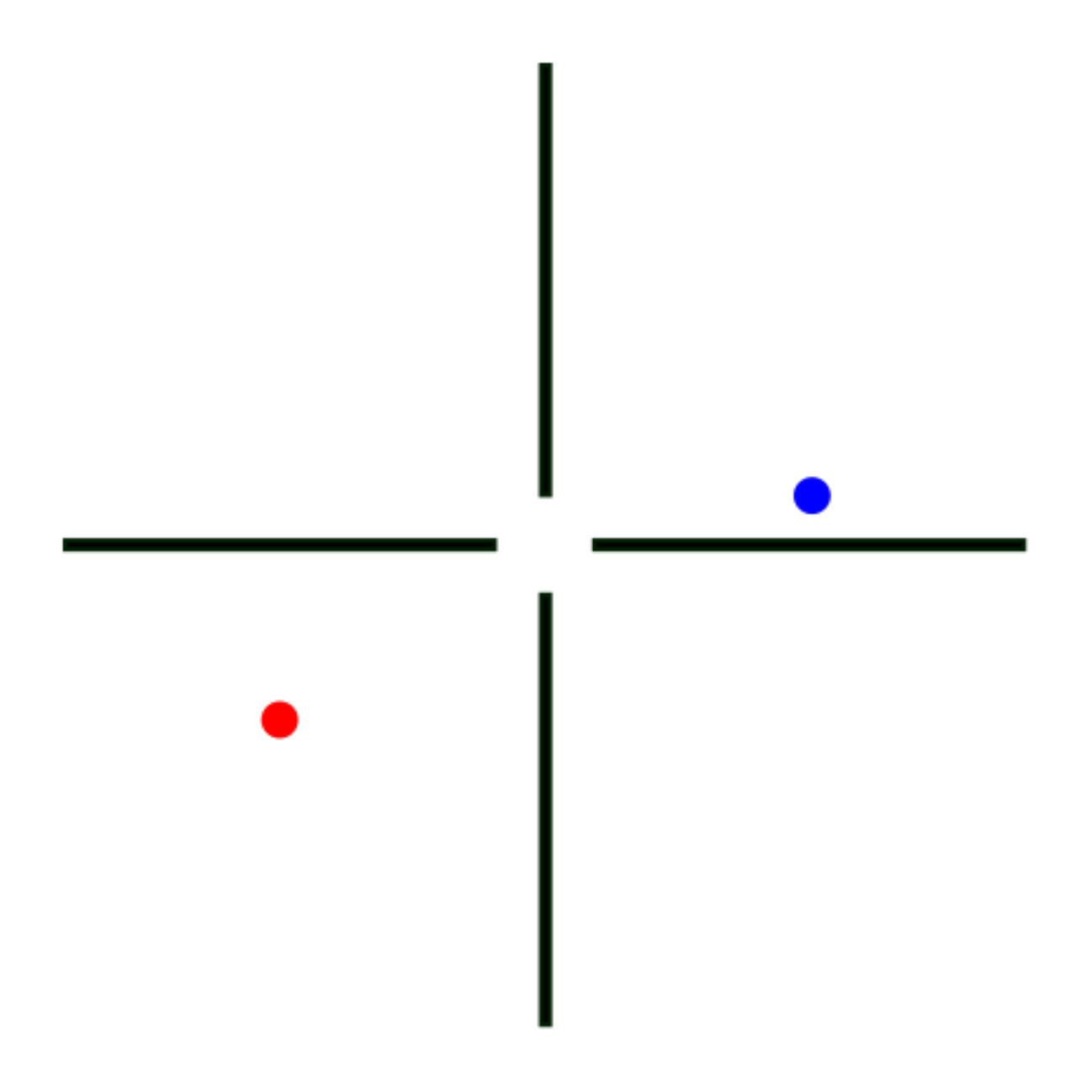} &
\includegraphics[height=0.165\textwidth]{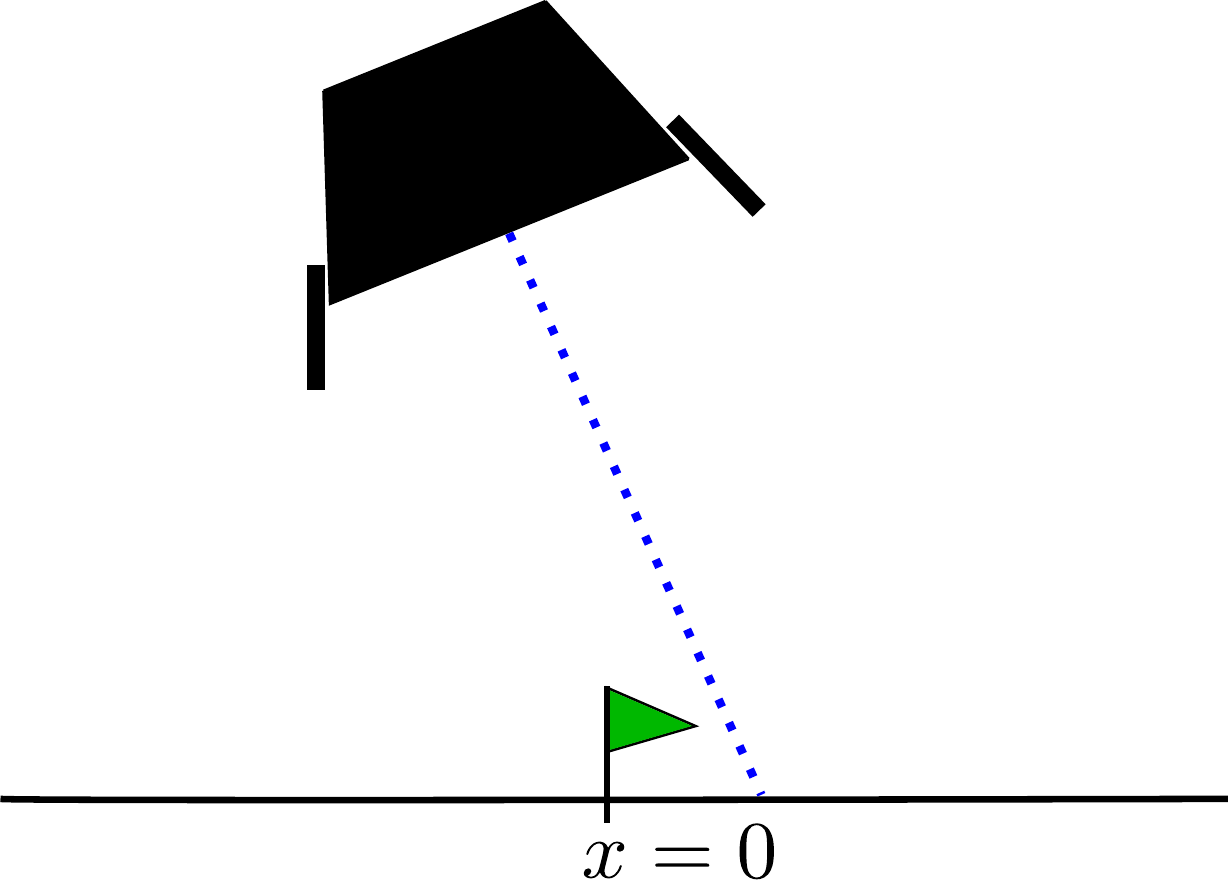} \\
(a) & (b) & (c) & (d) & (e)
\end{tabular}
\caption{Illustrations of (a) the Pushbox2D, (b) the Parking2D, (c) the SensorPlacement-8, (d) the VDP-Tag and (e) the LunarLander problems. Goal regions in are marked as green circles or a green flag.}
\label{f:problemScenarios}
\end{figure*}


\begin{table*}
\caption{Summary of the state, action and observation spaces for each problem scenario.}
\centering
\label{t:pomdp_properties}
\begin{tabular}{l|lll}
& \stSpace & \actSpace & \obsSpace \\ \hline \hline
Pushbox2D & $\mathbb{R}^{4}$ & $[-1, 1]^2$ & $24$ observations \\
Pushbox3D & $\mathbb{R}^{6}$ & $[-1, 1]^3$ & $288$ observations \\
Parking2D & $\mathbb{R}^{4}$ & $[-\frac{\pi}{2}, \frac{\pi}{2}]\times [-8, 8]$ & $4$ observations \\
Parking3D & $\mathbb{R}^{5}$ & $[-\frac{\pi}{2}, \frac{\pi}{2}]\times [-8, 8]\times [-\frac{7}{2}, \frac{7}{2}]$ & $4$ observations \\
SensorPlacement-D & $\mathbb{R}^{D}$ & $\mathbb{R}^D$ & $4$ observations \\
VDP-Tag & $\mathbb{R}^{4}$ & $[0, 2\pi)\times \{0, 1\}$ & $\mathbb{R}^8$ \\
LunarLander & $\mathbb{R}^{6}$ & $\mathbb{R}^+ \times \mathbb{R}$ & $\mathbb{R}^3$ \\
\end{tabular}
\end{table*}

\subsubsection{Pushbox}
Pushbox is a scalable motion planning problem proposed in\ccite{seiler2015online} which is motivated by air hockey. A disk-shaped robot \commNew{(blue disk in \Cref{f:problemScenarios}(a))} has to push a disk-shape puck \commNew{(red disk in \Cref{f:problemScenarios}(a))} into a goal region \commNew{(green circle in \Cref{f:problemScenarios}(a))} by bumping into it, while avoiding any collision of itself and the puck with a boundary region \commNew{(black region if \Cref{f:problemScenarios}(a)). The robot receives a reward of $1,000$ when the puck is pushed into the goal region, while it receives a penalty of $-1,000$ if the robot or the puck collides with the boundary region. Additionally, the robot receives a penalty of $-10$ for every step.} The robot can move freely in the environment by choosing a displacement vector. Upon bumping into the puck, the puck is pushed away and the motion of the puck is affected by noise. We consider two variants of the problem, \textbf{Pushbox2D} and \textbf{Pushbox3D} that differ in the dimensionality of the state and action spaces. For the \textbf{Pushbox2D} problem (illustrated in \Cref{f:problemScenarios}(a)), the robot and the puck operate on a 2D-plane, whereas for \textbf{Pushbox3D}, both the robot and the puck operate inside a 3D-environment. Let us first describe the \textbf{Pushbox2D} variant. \commNew{The state space consists of the $xy$-locations of both the robot and the puck, \ie, $\stSpace = \mathbb{R}^4$, while the action space is defined by $\actSpace = [-1, 1] \times [-1, 1]$. If the robot is not in contact with the puck during a move, the state evolves deterministically according to $f(\st, \act) = (x_r + a_x, y_r + a_y, x_p, y_p)^T$, where $(x_r, y_r)$ and $(x_p, y_p)$ are the $xy$-coordinates the robot and the puck respectively, corresponding to state \st, and $(\act_x, \act_y)$ is the displacement vector corresponding to action \act.} \commNew{In particular, if the robot bumps into the puck, the next position $(x_p', y_p')$ of the puck is computed as
\begin{equation}
\begin{pmatrix}
x_p'\\ 
y_p'
\end{pmatrix} = \begin{pmatrix}
x_p\\ 
y_p
\end{pmatrix} + 5r_s \left (\begin{pmatrix}
a_x\\ 
a_x
\end{pmatrix} \cdot \vec{n}\right )\left (\vec{n} + \begin{pmatrix}
r_x\\ 
r_y
\end{pmatrix} \right ),
\end{equation}
where the $``\cdot"$ operator denotes the dot product, $\vec{n}$ is the unit directional vector from the center of the robot to the center of the puck at the time of contact, and $r_s$ is a random variable drawn from a truncated Gaussian distribution $N(\mu, \sigma^2, l, u)$, which is the Gaussian distribution $N(\mu, \sigma^2)$ truncated to the interval $[l, u]$. For our experiments, we used $\mu=1.0$, $\sigma=0.1$, $l=\mu-\sigma = 0.9$ and $u=\mu + \sigma = 1.1$. The variables $r_x$ and $r_y$ are random variables drawn from a truncated Gaussian distribution $N(0.0, 0.1^2, -0.1, 0.1)$.}

\commNew{The initial position position of the robot is known and is set to $x_r=5.5$ and $y_r=9.5$ respectively.} The initial puck position, however, is uncertain. \commNew{Its initial $x_p$ and $y_p$ coordinates are drawn from a truncated Gaussian distribution $N(5.5, 2.0^2, 3.5, 7.5)$, but the robot has access to a noisy bearing sensor to localize the puck and a noise-free collision sensor which detects contacts between the robot and the puck. In particular, given a state $\st\in\stSpace$, an observation $(\obs_c, \obs_b)$ consists of a binary component $\obs_c$ which indicates whether or not a contact between the robot and the puck occured, and a discretized bearing component $\obs_b$ calculated as 

\begin{equation}\label{eq:pushbox_bearing}
\obs_b = \mathrm{floor}\left (\frac{\mathrm{atan2}(y_o - y_r, x_o - x_r) + r_o}{\pi/6} \right ),
\end{equation}
where $x_r, y_r$ and $x_o, y_o$ are the $xy$-coordinates of the robot and the puck corresponding to the state \st, and $r_o$ is a random angle (expressed in radians) drawn from a truncated Gaussian distribution $N(0.0, (\frac{\pi}{18})^2, -\frac{\pi}{18}, \frac{\pi}{18})$. Due to the $\mathrm{floor}$ operator in \cref{eq:pushbox_bearing}, the number of discretized bearing observation is 12, thus the observation space consists of 24 unique observations.

\textbf{Pushbox3D} is a straightforward extension of the \textbf{Pushbox2D} problem: The state space is defined as $\stSpace = \mathbb{R}^6$, consisting of the $xyz$-locations of the robot and the puck. The action space is $\actSpace = [-1, 1]\times[-1, 1]\times[-1, 1]$, where each $\act\in\actSpace$ describes a 3D-displacement vector of the robot. The transition dynamics are defined similarly to \textbf{Pushbox2D}, except that all quantities are computed in 3D. The observation space is extended with an additional bearing observation, computed according to \cref{eq:pushbox_bearing}, but with the term $\mathrm{atan2}(y_o - y_r, x_o - x_r)$ being replaced by $\mathrm{atan2}(z_o - z_r, y_o - y_r)$, where $z_r$ and $z_o$ are the $z$-coordinates of the robot and the puck respectively. The reward function is the same as in the \textbf{Pushbox2D} problem.

For both variants of the problem the discount factor is $\gamma=0.95$ and a run is considered successful if the robot manages to push the puck into a goal region within 50 steps, while avoiding collisions of itself and the puck with the boundary region.}


\subsubsection{Parking}
An autonomous vehicle with deterministic dynamics operates in a 3D-environment populated by obstacles, shown in \fref{f:problemScenarios}(b). The goal of the vehicle is to safely navigate to a goal area located between the obstacles while avoiding any collision with the obstacles (black regions in \fref{f:problemScenarios}(b)). \commNew{The vehicle receives a reward of $100$ when reaching the goal area, while a collisions with the obstaces are penalized by $-100$. Attionally, the vehicle receives a penalty of $-1$ for every step.}  We consider two variants of the problem, \textbf{Parking2D} and \textbf{Parking3D}. For \textbf{Parking2D}, the vehicle navigates on a 2D-plane, \commNew{whereas for \textbf{Parking3D}, the vehicle operates in 3D-space. We first describe the \textbf{Parking2D} variant: The state space is $\stSpace = \mathbb{R}^4$ and consists of the $xy$-position of the vehicle on the plane, its orientation $\theta$ and its velocity $v$.} The vehicle is controlled via a steering wheel angle $\act_{\theta}$ and acceleration $\act_{v}$, \ie, the action space is $\actSpace = \Omega\times\Phi$, where $\Omega = [-\frac{\pi}{2}, \frac{\pi}{2}]$ is the continuous set of steering wheel angles and $\Phi = [-8, 8]$ is the continuous set of accelerations. \commNew{We assume that for a given state $\st\in\stSpace$ and action $\act\in\actSpace$, the state of the vehicle evolves acoording to the following deterministic second-order discrete-time dynamics model:
\begin{equation}\label{eq:parking_dynamics}
f(x, y, \theta, v, \act_{\theta}, \act_{v}) = \begin{bmatrix}
x + v\cos(\theta) \Delta \\ 
y + v\sin(\theta) \Delta \\
\theta + a_{\theta} \Delta \\
v + a_v \Delta 
\end{bmatrix},
\end{equation}
where $x$, $y$, $\theta$ and $v$ are the 2D-position, orientation and velocity corresponding to state \st, $\Delta=\frac{1}{3}s$ is a fixed control duration, and $\act_{\theta}$, $\act_{v}$ are the steering wheel angle and acceleration components of the action. To compute the next state $\st'$, given $\st$ and \act, we numerically integrate \cref{eq:parking_dynamics} for 3 steps.}

There are three distinct areas in the environment, each consisting of a different type of terrain (colored areas in \fref{f:problemScenarios}(b)). Upon traversal, the vehicle receives an observation regarding the terrain type, which is only correct 70\% of the time due to sensor noise. \commNew{If the vehicle is outside the terrains, we assume that it deterministically receives a \texttt{NULL} observation.} Initially the vehicle starts near one of three possible starting locations (red areas in \fref{f:problemScenarios}(b)) with equal probability. The exact initial position of the vehicle along the horizontal $y$-axis is then drawn uniformly from $U[-0.175, 0.175]$ around the starting location. For \textbf{Parking3D} the vehicle operates in the full 3D space, and we have additional continuous state and action components that model the vehicles elevation and change in elevation respectively. \commNew{We assume that the elevation of the vehicle changes according to $z + \Delta \act_h$, where $z\in\mathbb{R}$ is the elevation component of the state, and $\act_h\in[-\frac{7}{2}, \frac{7}{2}]$ is the elevation-change component of the action.} The discount factor for both variants is $\gamma = 0.95$ \commNew{and a run is considered successful if the vehicle enters the goal area within 50 steps while avoiding collisions with the obstacles.}

Two properties make this problem challenging. First is the multi-modal beliefs which require the vehicle to traverse the different terrains for a sufficient amount of time to localize itself before attempting to reach the goal. \commNew{Second, due to the narrow passage that leads to the goal, small perturbations from the optimal action can quickly result in collisions with the obstacle. As a consequence, good rewards are sparse and a POMDP solver must discover them quickly in order to compute a near-optimal strategy.}

\subsubsection{SensorPlacement}
We propose a scalable motion planning under uncertainty problem in which a $D$-DOF manipulator with $D$ revolute joints operates in muddy water inside a 3D environment. The robot is located in front of a marine structure, represented as four distinct walls, and its task is to mount a sensor at a particular goal area between the walls (rewarded with 1,000) while having imperfect information regarding its exact joint configuration. To localize itself, the robot's end-effector is equipped with a touch sensor. Upon touching a wall, it provides noise-free information regarding \commNew{which wall is being touched, while we assume that the sensor deterministically outputs a \texttt{NULL} observation in a contact-free state}. However, in order to avoid damage, the robot must avoid collisions (penalized by $-500$) between any of its other links and the walls. The state space of the robot consists of the set of joint-angles for each joint. The action space is $\actSpace \subset \mathbb{R}^D$, where $\act\in\actSpace$ is a vector of joint velocities. Due to underwater currents, the robot is subject to random control errors \commNew{and the joint-angles corresponding to a state evolve according to
\begin{equation}
\theta' = \theta + \act + r,
\end{equation}
where $\theta$ is the set of joint angles corresponding to the current state, and $r$ is a random vector sampled from a multivariate Gaussian distribution $N({\bf 0}, \sigma^2 I)$, where $I$ is the identity matrix of size $D$, and $\sigma^2=10^{-3}$.} Initially the robot is uncertain regarding its exact joint angle configuration. We assume that the initial joint angles are distributed uniformly according to $U[\theta_l, \theta_u]$, where $\theta_l = \theta_0 - h$ and $\theta_u = \theta_0 + h$, with $\theta_0$ corresponding to the configuration where all joint angles are zero, except for the second and third joint whose joint angles are $-1.57$ and $1.57$ respectively and $h = (0.1, \ldots, 0.1) \in \mathbb{R}^{D}$ (units are in radians). We consider four variants of the problem, denoted as SensorPlacement-$D$, with $D\in\{6, 8, 10, 12\}$, that differ in the degrees-of-freedom (number of revolute joints and thus the dimensionality of the action space) of the manipulator. \fref{f:problemScenarios}(c) illustrates the SensorPlacement-8 problem, where the colored areas represent the walls and the green sphere represents the goal area. The discount factor is $\gamma=0.95$ \commNew{and a run is considered successful if the manipulator mounts the sensor within 50 steps while avoiding collisions with the walls}. To successfully mount the sensor at the target location, the robot must use its touch sensor to carefully reduce uncertainty regarding its configuration. This is challenging, since a slight variation in the performed actions can quickly lead to collisions with the walls.

\subsubsection{Van Der Pol Tag}
Van Der Pol Tag (VDP-Tag) is a benchmark problem introduced in\ccite{sunberg2018online} in which an agent \commNew{(blue particle in \Cref{f:problemScenarios}(d))} operates in a 2D-environment. The goal is to tag a moving target \commNew{(red particle in \Cref{f:problemScenarios}(d))} whose motion is described by the Van Der Pol differential equation and disturbed by Gaussian noise with standard deviation $\sigma=0.05$. Initially, the position of the target is unknown. The agent travels at a constant speed but can pick its direction of travel at each step and whether to activate a costly range sensor, \ie, the action space is $\actSpace = [0, 2\pi) \times \{0, 1\}$, where the first component is the direction of travel and the second component is the activation/deactivation of the range sensor. The robot receives observations from its range sensor via 8 beams (\ie, $\obsSpace = \mathbb{R}^8$) that measure the agent's distance to the target if the target is within the beam's range. These measurements are more accurate when the range sensor is active. While the target moves freely in the environment, the agent's movements are restricted by four obstacles in the environment, \commNew{shown in \Cref{f:problemScenarios}(d). Catching the target is reward by $100$, while activating the range sensor is penalized by $-5$. Additionally, each step incurs a penalty of $-1$. The discount factor is $\gamma=0.95$ and a run is considered successful of the agent catches the target within $50$ steps.} More details of the VDP-Tag problem can be found in\ccite{sunberg2018online}.

\commNew{Note that in this problem, the action space consists of two disconnected components, namely $[0, 2\pi)\times \{0\}$ and $[0, 2\pi)\times \{1\}$. To allow a Voronoi tree $\VT(\bel)$ in \solverAbbr to cover the entire action space, we ensure that once the root node of $\VT(\bel)$ is split, we set the range sensor component of the representative actions of the two resulting child cells to $0$ and $1$ respectively, such that one child cell covers the component $[0, 2\pi)\times \{0\}$, and the other child cell covers the component $[0, 2\pi)\times \{1\}$.}

\subsubsection{LunarLander}
\commNew{This problem is a partially observable adaptation of Atari's Lunar Lander game. 
For this problem, the state, action and observation spaces are all continuous. The objective is to control a lander vehicle to safely land at a target zone located on the moon's surface. The lander operates on a $xy$-plane and its state is a 6D-vector $(x, y, \theta, \dot{x}, \dot{y}, \dot{\theta})$, where $x\in\mathbb{R}$ and $y\in\mathbb{R}$ are the horizontal and vertical positions of the lander and $\theta\in\mathbb{R}$ its orientation. $\dot{x}\in\mathbb{R}$, $\dot{y}\in\mathbb{R}$ and $\dot{\theta}\in\mathbb{R}$ represent the lander's horizontal, vertical and angular velocities respectively. The action space is $\actSpace=\Lambda\times \Psi$, where $\Lambda = \mathbb{R}^+$ is the set of linear accelerations along the lander's vertical axis, and $\Psi = \mathbb{R}$ is the set of angular accelerations about the lander's geometric center. The initial belief of the lander is a multivariate Gaussian distribution with mean $\mu = (0, 10, 0, 0, -10, 0)^T$ and covariance matrix $\Sigma = \mathrm{diag}(1.5^2, 1.0^2, 0.1^2, 0^2, 0.5^2, 0.1^2)$.

We assume that the state of the lander evolves according to the following second-order discrete-time stochastic model:
\begin{equation}\label{eq:landerDynamics}
f(x, y, \theta, \dot{x}, \dot{y}, \dot{\theta}, \tilde{\lambda}, \tilde{\psi}) = \begin{bmatrix}
x + \dot{x} \Delta  \\
y +  \dot{y} \Delta \\
\theta + {\dot\theta} \Delta\\ 
\dot{x} + (-\tilde{\lambda}\sin(\theta)M)\Delta \\ 
\dot{y} + (\tilde{\lambda}\cos(\theta)M - G)\Delta \\
\dot{\theta} + H \tilde{\psi}\Delta 
\end{bmatrix},
\end{equation}
where $\tilde{\lambda} = \lambda + e_\lambda$ and $\tilde{\psi} = \psi + e_\psi$, with $\lambda$ and $\psi$ being the vertical and angular accelerations corresponding to the action, and $e_\lambda$ and $e_\psi$ are random control errors drawn from zero-mean Gaussian distributions with standard deviations $\sigma_\lambda = 1\times 10^{-4}$ and $\sigma_\psi = 5\times 10^{-2}$ respectively. The variable $\Delta = 0.2s$ is a constant step size, whereas $F=40$ and $H=2$ are motor constants. The variable $G = -9.81m/s^2$ is a constant describing the gravitational acceleration along the $y$-axis.
To obtain the next state, given the current state and an action, we numerically integrate the system in \cref{eq:landerDynamics} for 5 steps.

The lander perceives information regarding its state via three noisy sensors: The first two sensors measure the lander's horizontal and angular velocities, whereas the third sensor measures the distance to the ground along the lander's vertical axis (dashed line in \Cref{f:problemScenarios}(e)). The readings of all three sensors are disturbed by standard-Gaussian noise.

The reward function is defined by
\begin{equation}\label{eq:landerReward}
r_t = 
\begin{cases}
    -1000,& \text{if } \theta\geq 0.5\ \text{or } y_t < 0\\
    100 - \left |x_t \right | - \left |\theta_t \right | - \dot{y}_t^{2},& \text{if } y_t\leq 0.3\\
    -1,              & \text{otherwise.}
\end{cases}
\end{equation}

The first term in \cref{eq:landerReward} encourages the lander to prevent dangerous angles and crashing into the ground. The second term encourages the lander to safely land at $x = 0$ with an upright angle and a small vertical velocity. The third term encourages the lander to land as quickly as possible. The discount factor is $\gamma=0.95$ and a run is considered successful if the lander's vertical position reaches $y \leq 0.3$ within $50$ steps, without crashing into the ground ($y< 0$).

}

\subsection{Experimental Setup}
 
\commNew{The purpose of our experiments is three-fold: First is to evaluate \solverAbbr and compare it with two state-of-the-art online POMDP solvers for continuous actions spaces, POMCPOW\ccite{sunberg2018online} and VOMCPOW\ccite{lim2020voronoi}. The results of those experiments are discussed in \Cref{sssec:comparisonWithBaseline}.

Second is to investigate the importance of the different components of \solverAbbr, specifically the Voronoi tree-based partitioning, the cell-diameter-aware exploration term in \cref{eq:u_value}, and the stochastic Bellman backups. For this purpose, we implemented the original \solverAbbr and three modifications. First is \textbf{\solverAbbr-R}, which replaces the Voronoi decomposition of \solverAbbr with a simple rectangular-based method: Each cell in the partition is a hyper-rectangle that is subdivided by cutting it in the middle along the longest side (with ties broken arbitrarily). The second variant is  \textbf{\solverAbbr(L=0)}, which is \solverAbbr where \cref{eq:u_value} reduces to the standard UCB1 bound. For the third variant, \textbf{\solverAbbr-MC}, we replace the stochastic Bellman backup in \cref{eq:\bellmanBackup} with the Monte Carlo backup in \cref{eq:mcBackup}, as used by POMCPOW and VOMCPOW. The results for this study are discussed in \Cref{sssec:effectsOfComponents}

Third is to test the effects of two algorithmic components of ADVT when applied to the baselines VOMCPOW and POMCPOW: In the original implementation of VOMCPOW and POMCPOW provided by the authors, the policy is recomputed after every planning step using a new search tree. In contrast, for discrete observation spaces, \solverAbbr applies ABT's\ccite{kurniawati2016online} strategy that re-uses the partial search tree (starting from the updated belief) constructed in the previous planning steps and improves the policy within the partial search tree. Therefore, for problems with discrete observation spaces, we also tested modified versions of VOMCPOW and POMCPOW, where we follow \solverAbbr's strategy of re-using the partial search trees. Note that for the VDP-Tag and LunarLander problems, we did not test the variants of VOMCPOW and POMCPOW that re-use partial search trees, since each observation that the agent perceives from the environment leads to a new search tree due to the continuous observation spaces. Moreover, to test the effects of \solverAbbr's stochastic Bellman backup strategy further, we implemented variants of VOMCPOW and POMCPOW to use stochastic Bellman backups instead of Monte Carlo backups. The results are discussed in \Cref{sssec:abaltionStudy}.}

To approximately determine the best parameters for each solver and problem, we ran a set of systematic preliminary trials by performing a grid-search over the parameter space. For each solver and problem, we used the best parameters and ran 1,000 simulation runs, with a fixed planning time of 1s CPU time for each solver and scenario. Each tested solver and the scenarios were implemented in C++ using the OPPT-framework\ccite{hoerger2018software}. All simulations were run single-threaded on an Intel Xeon Platinum 8274 CPU with 3.2GHz and 4GB of memory. 


\subsection{Results}
\begin{table*}[hbt!]
\centering
\caption{Average total discounted rewards and 95\% confidence intervals of ADVT, VOMCPOW and POMCPOW on the Pushbox, Parking, VDP-Tag and LunarLander problems. The average is taken over 1000 simulation runs per solver and problem, with a planning time of 1s per step. The best result for each problem is highlighted in bold.}\label{t:results_1}
\renewcommand{\arraystretch}{1.0}
\setlength{\tabcolsep}{2pt}
\begin{tabular}{l|*{6}{>{\hspace{0.55em}}rcl}}
& \multicolumn{3}{c}{Pushbox2D} & \multicolumn{3}{c}{Pushbox3D} &
	\multicolumn{3}{c}{Parking2D} & \multicolumn{3}{c}{Parking3D} & 
	\multicolumn{3}{c}{VDP-Tag} & \multicolumn{3}{c}{LunarLander} \\ \hline \hline
\solverAbbr & $\mathbf{351.6}$ & $\hspace{-0.2em}\pm\hspace{-0.2em}$ & $\mathbf{10.0}$ & $\mathbf{322.1}$ & $\hspace{-0.2em}\pm\hspace{-0.2em}$ &
$\mathbf{14.9}$ & $\mathbf{35.2}$ & $\hspace{-0.2em}\pm\hspace{-0.2em}$ & $\mathbf{1.9}$ & $\mathbf{32.6}$ & $\hspace{-0.2em}\pm\hspace{-0.2em}$ & $\mathbf{3.5}$ & $30.5$ & $\hspace{-0.2em}\pm\hspace{-0.2em}$ & $1.0$ & $\mathbf{26.01}$ & $\hspace{-0.2em}\pm\hspace{-0.2em}$ & $\mathbf{1.2}$ \\ 
VOMCPOW & $129.8$ & $\hspace{-0.2em}\pm\hspace{-0.2em}$ & $13.3$ & $73.5$ & $\hspace{-0.2em}\pm\hspace{-0.2em}$ & $13.8$ & $-0.78$ & $\hspace{-0.2em}\pm\hspace{-0.2em}$  &
$2.8$ & $-18.4$ & $\hspace{-0.2em}\pm\hspace{-0.2em}$ & $1.4$ & $\mathbf{32.9}$ & $\hspace{-0.2em}\pm\hspace{-0.2em}$ & $\mathbf{0.9}$ & $13.3$ & $\hspace{-0.2em}\pm\hspace{-0.2em}$ & $2.2$ \\
POMCPOW & $82.1$ & $\hspace{-0.2em}\pm\hspace{-0.2em}$ & $14.2$ & $3.6$ & $\hspace{-0.2em}\pm\hspace{-0.2em}$ & $12.9$ & $-5.1$ & $\hspace{-0.2em}\pm\hspace{-0.2em}$ & $3.0$ & $-25.7$ & $\hspace{-0.2em}\pm\hspace{-0.2em}$ & $1.4$ & $28.2$ & $\hspace{-0.2em}\pm\hspace{-0.2em}$ & $1.1$ & $13.5$ & $\hspace{-0.2em}\pm\hspace{-0.2em}$ & $2.1$ \\ 
\end{tabular}
\end{table*}

\begin{figure*}[hbt!]
\centering
\small
\includegraphics[width=0.75\textwidth]{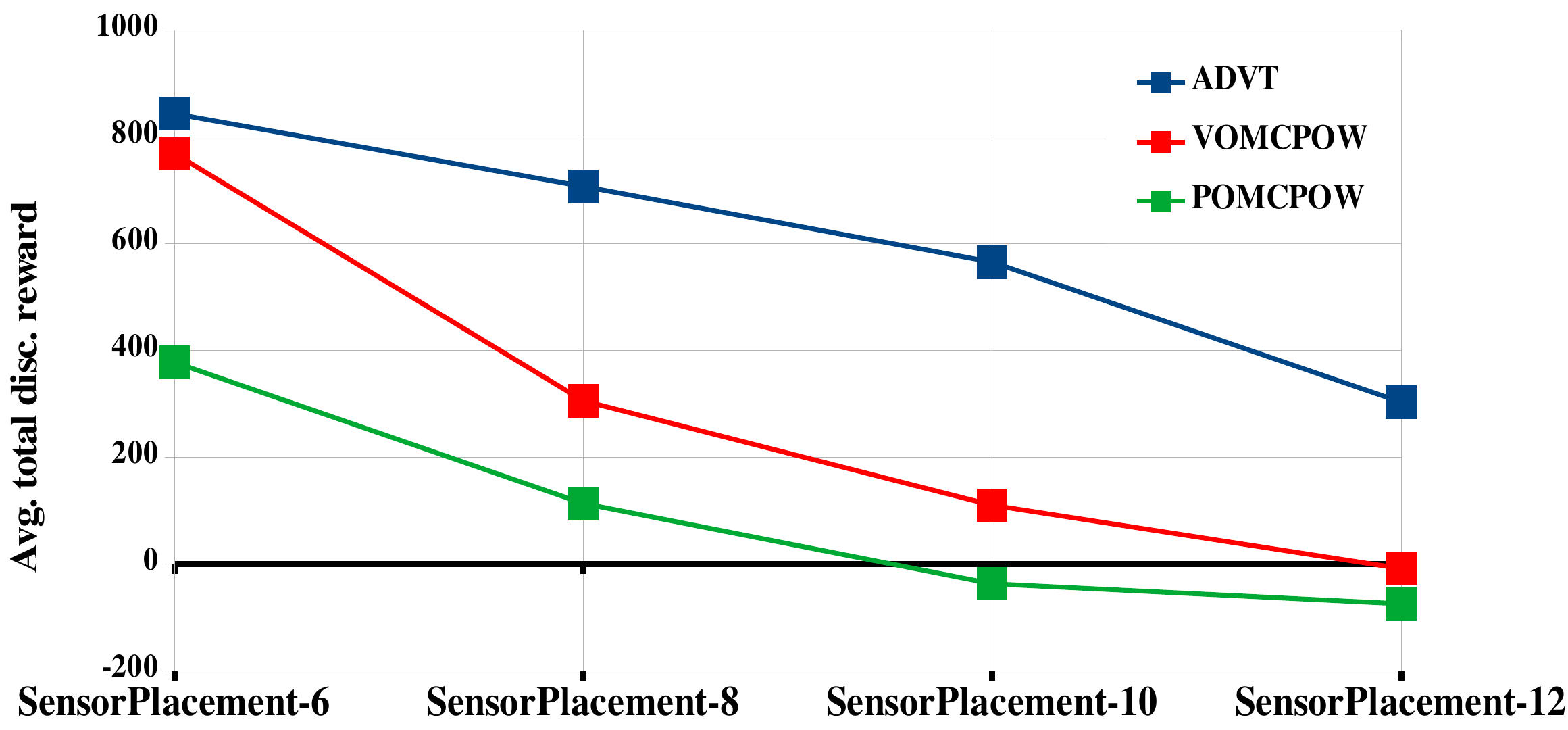}
\caption{Average total discounted rewards of all tested solvers on the SensorPlacement problems. The average is taken over 1,000 simulation runs per solver and problem.}
\label{f:ResultsSensorPlacement}
\end{figure*}

\begin{table*}[hbt!]
\centering
\caption{Average total discounted rewards and 95\% confidence intervals of all tested solvers on the Pushbox, Parking, VDP-Tag and LunarLander problems. The average is taken over 1000 simulation runs per solver and problem, with a planning time of 1s per step.}\label{t:results_ablation}
\renewcommand{\arraystretch}{1.0}
\setlength{\tabcolsep}{2pt}
\begin{tabular}{l|*{6}{>{\hspace{0.55em}}rcl}}
& \multicolumn{3}{c}{Pushbox2D} & \multicolumn{3}{c}{Pushbox3D} &
	\multicolumn{3}{c}{Parking2D} & \multicolumn{3}{c}{Parking3D} & 
	\multicolumn{3}{c}{VDP-Tag} & \multicolumn{3}{c}{LunarLander} \\ \hline \hline
\solverAbbr-R & $371.4$ & $\hspace{-0.2em}\pm\hspace{-0.2em}$ & $9.8$ & $321.2$ & $\hspace{-0.2em}\pm\hspace{-0.2em}$ & $15.1$ &
$38.9$ & $\hspace{-0.2em}\pm\hspace{-0.2em}$ & ${1.8}$ & $24.3$ & $\hspace{-0.2em}\pm\hspace{-0.2em}$ & $3.4$ & $30.2$ & $\hspace{-0.2em}\pm\hspace{-0.2em}$ & $1.0$ & $29.6$ & $\hspace{-0.2em}\pm\hspace{-0.2em}$ & $1.1$ \\
\solverAbbr(L=0) & $340.8$ & $\hspace{-0.2em}\pm\hspace{-0.2em}$ & $14.7$ & $294.6$ & $\hspace{-0.2em}\pm\hspace{-0.2em}$ & $13.3$ & $29.2$ & $\hspace{-0.2em}\pm\hspace{-0.2em}$ & $3.5$ & $18.6$ & $\hspace{-0.2em}\pm\hspace{-0.2em}$ & $1.7$ & $28.7$ & $\hspace{-0.2em}\pm\hspace{-0.2em}$ & $1.1$ & $24.7$ & $\hspace{-0.2em}\pm\hspace{-0.2em}$ & $1.2$\\
\solverAbbr-MC & $319.6$ & $\hspace{-0.2em}\pm\hspace{-0.2em}$ & $13.7$ & $311.0$ & $\hspace{-0.2em}\pm\hspace{-0.2em}$ & $16.2$ & $-3.2$ & $\hspace{-0.2em}\pm\hspace{-0.2em}$ & $1.8$ & $-14.7$ & $\hspace{-0.2em}\pm\hspace{-0.2em}$ & $0.5$ & $33.5$ & $\hspace{-0.2em}\pm\hspace{-0.2em}$ & $0.8$ & $21.9$ & $\hspace{-0.2em}\pm\hspace{-0.2em}$ & $1.7$ \\ \hline
VOMCPOW+our \reuseTree+our \bellmanBackup & $322.9$ & $\hspace{-0.2em}\pm\hspace{-0.2em}$ & $12.1$ & $274.9$ & $\hspace{-0.2em}\pm\hspace{-0.2em}$ & $14.2$ & $28.2$ & $\hspace{-0.2em}\pm\hspace{-0.2em}$ & $1.8$ & $24.4$ & $\hspace{-0.2em}\pm\hspace{-0.2em}$ & $2.4$ & - &  &  & -\\
VOMCPOW+our \bellmanBackup & $316.3$ & $\hspace{-0.2em}\pm\hspace{-0.2em}$ & $13.6$ & $134.2$ & $\hspace{-0.2em}\pm\hspace{-0.2em}$ & $17.4$ & $27.5$ & $\hspace{-0.2em}\pm\hspace{-0.2em}$ & $1.9$ & $23.7$ & $\hspace{-0.2em}\pm\hspace{-0.2em}$ & $2.5$ & $29.9$ & $\hspace{-0.2em}\pm\hspace{-0.2em}$ & $1.0$ & $19.9$ & $\hspace{-0.2em}\pm\hspace{-0.2em}$ & $1.6$ \\
VOMCPOW+our \reuseTree & $316.0$ & $\hspace{-0.2em}\pm\hspace{-0.2em}$ & $12.3$ & $268.9$ & $\hspace{-0.2em}\pm\hspace{-0.2em}$ & $14.2$ & $-0.42$ & $\hspace{-0.2em}\pm\hspace{-0.2em}$ & $2.8$ & $-15.7$ & $\hspace{-0.2em}\pm\hspace{-0.2em}$ & $1.5$ & - &  &  & - \\ \hline
POMCPOW+our \reuseTree+our \bellmanBackup & $314.2$ & $\hspace{-0.2em}\pm\hspace{-0.2em}$ & $13.0$ & $245.7$ & $\hspace{-0.2em}\pm\hspace{-0.2em}$ & $14.1$ & $27.7$ & $\hspace{-0.2em}\pm\hspace{-0.2em}$ & $1.8$ & $8.8$ & $\hspace{-0.2em}\pm\hspace{-0.2em}$ & $2.6$ & - &  &  & -\\ 
POMCPOW+our \bellmanBackup & $300.6$ & $\hspace{-0.2em}\pm\hspace{-0.2em}$ & $12.6$ & $128.8$ & $\hspace{-0.2em}\pm\hspace{-0.2em}$ & $17.5$ & $24.2$ & $\hspace{-0.2em}\pm\hspace{-0.2em}$ & $1.9$ & $-10.4$ & $\hspace{-0.2em}\pm\hspace{-0.2em}$ & $2.1$ & $27.5$ & $\hspace{-0.2em}\pm\hspace{-0.2em}$ & $1.2$ & $18.4$ & $\hspace{-0.2em}\pm\hspace{-0.2em}$ & $2.2$\\ 
POMCPOW+our \reuseTree & $270.6$ & $\hspace{-0.2em}\pm\hspace{-0.2em}$ & $18.9$ & $203.7$ & $\hspace{-0.2em}\pm\hspace{-0.2em}$ & $14.3$ & $-5.2$ & $\hspace{-0.2em}\pm\hspace{-0.2em}$ & $2.9$ & $-22.8$ & $\hspace{-0.2em}\pm\hspace{-0.2em}$ & $1.3$ & - &  &  & -\\
\end{tabular}
\end{table*}

\begin{table*}[hbt!]
\caption{Average total discounted rewards and 95\% confidence intervals of all tested solvers on the SensorPlacement problems. The average is taken over 1000 simulation runs per solver and problem, with a planning time of 1s per step.}\label{t:results_2}
\centering
\begin{adjustbox}{max width=\textwidth}
\setlength{\tabcolsep}{2pt}
\normalsize
\begin{tabular}{l|*{4}{>{\hspace{0.55em}}rcl}}
& \multicolumn{3}{c}{SensorPlacement-6} & \multicolumn{3}{c}{SensorPlacement-8} & \multicolumn{3}{c}{SensorPlacement-10} & \multicolumn{3}{c}{SensorPlacement-12} \\ \hline \hline
\solverAbbr & $\mathbf{842.8}$ & $\hspace{-0.2em}\pm\hspace{-0.2em}$ & $\mathbf{9.5}$ & $\mathbf{706.8}$ & $\hspace{-0.2em}\pm\hspace{-0.2em}$ & $\mathbf{17.5}$ & $\mathbf{565.1}$ & $\hspace{-0.2em}\pm\hspace{-0.2em}$ & $\mathbf{21.7}$ & $\mathbf{303.0}$ & $\hspace{-0.2em}\pm\hspace{-0.2em}$ & $\mathbf{19.8}$ \\
\solverAbbr-R & $676.3$ & $\hspace{-0.2em}\pm\hspace{-0.2em}$ & ${19.7}$ & $238.1$ & $\hspace{-0.2em}\pm\hspace{-0.2em}$ & $33.5$ & $28.7$ & $\hspace{-0.2em}\pm\hspace{-0.2em}$ & $18.4$ & $-17.3$ & $\hspace{-0.2em}\pm\hspace{-0.2em}$ & $7.4$ \\
\solverAbbr ($L=0$) & $780.4$ & $\hspace{-0.2em}\pm\hspace{-0.2em}$ & $12.6$ & $448.8$ & $\hspace{-0.2em}\pm\hspace{-0.2em}$ & $15.9$ & $325.3$ & $\hspace{-0.2em}\pm\hspace{-0.2em}$ & $16.2$ & $102.5$ & $\hspace{-0.2em}\pm\hspace{-0.2em}$ & $7.4$ \\
\solverAbbr-MC & $812.6$ & $\hspace{-0.2em}\pm\hspace{-0.2em}$ & $11.4$ & $692.7$ & $\hspace{-0.2em}\pm\hspace{-0.2em}$ & $17.7$ & $551.2$ & $\hspace{-0.2em}\pm\hspace{-0.2em}$ & $18.3$ & $293.5$ & $\hspace{-0.2em}\pm\hspace{-0.2em}$ & $19.6$ \\ \hline
VOMCPOW & $768.5$ & $\hspace{-0.2em}\pm\hspace{-0.2em}$ & $16.4$ & $305.6$ & $\hspace{-0.2em}\pm\hspace{-0.2em}$ & $25.8$ & $110.1$ & $\hspace{-0.2em}\pm\hspace{-0.2em}$ & $24.5$ & $-8.2$ & $\hspace{-0.2em}\pm\hspace{-0.2em}$ & $13.2$\\
VOMCPOW+our \reuseTree+our \bellmanBackup & $823.4$ & $\hspace{-0.2em}\pm\hspace{-0.2em}$ & $15.1$ & $679.1$ & $\hspace{-0.2em}\pm\hspace{-0.2em}$ & $17.9$ & $481.6$ & $\hspace{-0.2em}\pm\hspace{-0.2em}$ & $22.2$ & $191.3$ & $\hspace{-0.2em}\pm\hspace{-0.2em}$ & $17.6$ \\
VOMCPOW+our \bellmanBackup & $779.2$ & $\hspace{-0.2em}\pm\hspace{-0.2em}$ & $16.1$ & $654.9$ & $\hspace{-0.2em}\pm\hspace{-0.2em}$ & $16.3$ & $453.6$ & $\hspace{-0.2em}\pm\hspace{-0.2em}$ & $22.8$ & $182.4$ & $\hspace{-0.2em}\pm\hspace{-0.2em}$ & $17.7$ \\
VOMCPOW+our \reuseTree & $817.2$ & $\hspace{-0.2em}\pm\hspace{-0.2em}$ & $15.8$ & $663.4$ & $\hspace{-0.2em}\pm\hspace{-0.2em}$ & $18.6$ & $476.0$ & $\hspace{-0.2em}\pm\hspace{-0.2em}$ & $22.7$ & $189.9$ & $\hspace{-0.2em}\pm\hspace{-0.2em}$ & $18.0$ \\ \hline
POMCPOW & $377.6$ & $\hspace{-0.2em}\pm\hspace{-0.2em}$ & $23.5$ & $113.4$ & $\hspace{-0.2em}\pm\hspace{-0.2em}$ & $24.2$ & $-36.8$ & $\hspace{-0.2em}\pm\hspace{-0.2em}$ & $11.3$ & $-74.3$ & $\hspace{-0.2em}\pm\hspace{-0.2em}$ & $12.9$ \\
POMCPOW+our \reuseTree+our \bellmanBackup & $659.3$ & $\hspace{-0.2em}\pm\hspace{-0.2em}$ & $17.2$ & $428.7$ & $\hspace{-0.2em}\pm\hspace{-0.2em}$ & $21.5$ & $114.6$ & $\hspace{-0.2em}\pm\hspace{-0.2em}$ & $16.6$ & $-1.9$ & $\hspace{-0.2em}\pm\hspace{-0.2em}$ & $6.6$ \\
POMCPOW+our \bellmanBackup & $562.5$ & $\hspace{-0.2em}\pm\hspace{-0.2em}$ & $18.3$ & $408.7$ & $\hspace{-0.2em}\pm\hspace{-0.2em}$ & $19.1$ & $102.6$ & $\hspace{-0.2em}\pm\hspace{-0.2em}$ & $14.4$ & $-6.5$ & $\hspace{-0.2em}\pm\hspace{-0.2em}$ & $7.1$ \\
POMCPOW+our \reuseTree & $653.2$ & $\hspace{-0.2em}\pm\hspace{-0.2em}$ & $17.3$ & $425.2$ & $\hspace{-0.2em}\pm\hspace{-0.2em}$ & $21.8$ & $111.3$ & $\hspace{-0.2em}\pm\hspace{-0.2em}$ & $16.8$ & $-2.1$ & $\hspace{-0.2em}\pm\hspace{-0.2em}$ & 6.8 \\ 
\end{tabular}
\end{adjustbox}
\end{table*}

\begin{table*}[hbt!]
\caption{Success rates of all tested solvers on the Pushbox, Parking, VDP-Tag and LunarLander problems. The success rate is with respect to 1,000 simulation per solver and problem, with a planning time of 1s per step.}\label{t:results_success_rate}
\centering
\renewcommand{\arraystretch}{1.0}
\setlength{\tabcolsep}{2pt}
\begin{tabular}{l|*{6}{>{\hspace{0.55em}}rcl}}
& \multicolumn{3}{c}{Pushbox2D} & \multicolumn{3}{c}{Pushbox3D} &
	\multicolumn{3}{c}{Parking2D} & \multicolumn{3}{c}{Parking3D} & 
	\multicolumn{3}{c}{VDP-Tag} & \multicolumn{3}{c}{LunarLander} \\ \hline \hline
\solverAbbr & $0.985$ &  &  & $0.969$ &  &  & $0.912$ &  &  & $0.916$ &  &  & $0.985$ & & & $0.969$ & & \\
\solverAbbr-R & $0.987$ &  &  & $0.968$ &  &  & $0.943$ &  &  & $0.906$ & &  & $0.984$ & & & $0.970$ & &  \\
\solverAbbr(L=0) & $0.966$ &  &  & $0.965$ &  &  & $0.857$ &  &  & $0.898$ &  &  & $0.980$ & & & $0.931$ & & \\
\solverAbbr-MC & $0.989$ &  &  & $0.972$ &  &  & $0.417$ &  &  & $0.337$ &  & & $0.991$ & & & $0.957$ & & \\ \hline
VOMCPOW & $0.754$ &  &  & $0.815$ &  &  & $0.512$ &  &  & $0.297$ & &  & $0.987$ & & & $0.859$ & &  \\
VOMCPOW+our \reuseTree+our \bellmanBackup & $0.985$ & &  & $0.970$ &  &  & $0.885$ &  &  & $0.886$ &  &  & -\\
VOMCPOW+our \bellmanBackup & $0.976$ & &  & $0.893$ &  &  & $0.882$ &  &  & $0.885$ &  &  & $0.986$ & & & $0.958$ & & \\
VOMCPOW+our \reuseTree & $0.975$ &  & & $0.939$ &  &  & $0.597$ &  &  & $0.314$ &  &  &- \\ \hline
POMCPOW & $0.712$ &  &  & $0.692$ &  & & $0.401$ & & & $0.125$ &  &  & $0.979$ & & & $0.876$ & & \\ 
POMCPOW+our \reuseTree+our \bellmanBackup & $0.974$ &  &  & $0.953$ &  &  & $0.853$ &  &  & $0.534$ &  &  & -\\
POMCPOW+our \bellmanBackup & $0.970$ &  &  & $0.891$ &  & & $0.793$ & & & $0.246$ &  &  & $0.976$ & & & $0.943$ & & \\ 
POMCPOW+our \reuseTree & $0.963$ &  &  & $0.969$ &  &  & $0.409$ &  &  & $0.122$ &  &  & -\\
\end{tabular}
\end{table*}

\begin{table*}[hbt!]
\caption{Success rates of all tested solvers on the SensorPlacement problems. The success rate is with respect to 1,000 simulation per solver and problem, with a planning time of 1s per step.}
\centering
\label{t:results_success_rate_2}
\begin{adjustbox}{max width=\textwidth}
\begin{tabular}{l|cccc}
& SensorPlacement-6 & SensorPlacement-8 & SensorPlacement-10 & SensorPlacement-12\\ \hline \hline
\solverAbbr & $0.981$ & $0.962$ & $0.834$ & $0.724$ \\
\solverAbbr-R & $0.832$ & $0.692$ & $0.756$ & $0.557$ \\
\solverAbbr ($L=0$) & $0.937$ & $0.726$ & $0.791$ & $0.601$ \\
\solverAbbr-MC & $0.964$ & $0.959$ & $0.828$ & $0.719$ \\ \hline
VOMCPOW & $0.923$ & $0.721$ & $0.645$ & $0.583$\\ 
VOMCPOW+our \reuseTree+our \bellmanBackup & $0.979$ & $0.951$ & $0.807$ & $0.703$ \\ 
VOMCPOW+our \bellmanBackup & $0.924$ & $0.883$ & $0.798$ & $0.702$ \\
VOMCPOW+our \reuseTree & $0.967$ & $0.891$ & $0.803$ & $0.698$ \\ \hline
POMCPOW & $0.738$ & $0.636$ & $0.519$ & $0.321$ \\
POMCPOW+our \reuseTree+our \bellmanBackup & $0.829$ & $0.794$ & $0.646$ & $0.575$ \\
POMCPOW+our \bellmanBackup & $0.794$ & $0.779$ & $0.641$ & $0.563$ \\
POMCPOW+our \reuseTree & $0.826$ & $0.781$ & $0.657$ & $0.578$ \\
\end{tabular} 
\end{adjustbox}
\end{table*}

\subsubsection{Comparison with State-of-the-Art Methods}\label{sssec:comparisonWithBaseline}
\tref{t:results_1} shows the average total discounted rewards of all tested solvers on the Pushbox, Parking, VDP-Tag and LunarLander problems, while \fref{f:ResultsSensorPlacement} shows the results for the SensorPlacement problems. Detailed results for the SensorPlacement problems are presented in \Cref{t:results_2} while results on the success rates of the tested solvers are shown in \Cref{t:results_success_rate} and \Cref{t:results_success_rate_2}. 

\solverAbbr generally outperforms all other methods, except for VDP-Tag, where VOMCPOW performs better. Interestingly, as we will see in \Cref{sssec:effectsOfComponents}, the variant of \solverAbbr that uses Monte-Carlo backups instead of stochastic Bellman backups (\solverAbbr-MC) outperforms all other methods in this problem, which supports \solverAbbr's effectiveness in handling continuous actions. 
The results for the SensorPlacement problems indicate that \solverAbbr scales well to higher-dimensional action spaces. \commNew{Additionally, the results on the VDP-Tag and LunarLander problems indicate that \solverAbbr is capable of handling continuous observation spaces well.}

\solverAbbr performs well in terms of the success rate, too (\Cref{t:results_success_rate} and \Cref{t:results_success_rate_2}). \solverAbbr maintains more than $90$\% success rate in the Pushbox, Parking, VDP-Tag and LunarLander problems. \commNew{In the Parking3D problem, VOMCPOW's and POMCPOW's success rate can be as low as $\sim 30$\% and $12.5$\%.} \commNew{Similarly, in the SensorPlacement problems \solverAbbr achieves a higher success rate compared to VOMCPOW and POMCPOW. While in the SensorPlacement-6 problem, the gap between \solverAbbr and VOMCPOW is relatively small ($\sim 98\%$ for \solverAbbr and $\sim 92\%$ for VOMCPOW), the gap increases as the dimensionality of the action space increases. In the SensorPlacement-12 problem, \solverAbbr achieves a success rate of $>72\%$, while VOMCPOW's success rate drops to $<60\%$. POMCPOW achieves a success rate of $<74\%$ in the SensorPlacement-6 problem, while only achieving a success rate of $<32\%$ in the SensorPlacement-12 problem.} 

\subsubsection{Understanding the Effects of Different Components of \solverAbbr.}\label{sssec:effectsOfComponents}

\commNew{In this section we present results demonstrating the importance of three key algorithmic components of ADVT, namely the Voronoi-based partitions, the cell-size-aware optimistic upper-confidence bound and the Stochastic Bellman backups.}

\paragraph{Effects of Voronoi-based partitioning for \solverAbbr.}
To understand the benefit of our Voronoi-based partitioning method, we compare the results of \solverAbbr with those of \solverAbbr-R. \Cref{t:results_ablation} shows that \solverAbbr-R slightly outperforms \solverAbbr in the Pushbox2D, Parking2D, VDP-Tag and LunarLander problems, indicating that a rectangular-based partitioning works well for low-dimensional action spaces. However, \Cref{t:results_2} shows that \solverAbbr-R is uncompetitive in the SensorPlacement problems as the dimensionality of the action space increases. For rectangular-based partitionings, the diameters of the cells can shrink very slowly in higher-dimensional action spaces. Additionally, the cell refinement method is independent of the spatial locations of the sampled actions. Both properties result in loose optimistic upper-confidence bounds of the $Q$-values, leading to excessive exploration of sub-optimal areas of the action space. For Voronoi trees, the geometries (and therefore the diameters) of the cells are much more adaptive to the spatial locations of the sampled actions, leading to more accurate optimistic upper-confidence bounds of the associated $Q$-values which avoids over-exploration of areas in the action space that already contain sufficiently many sampled actions. 

\paragraph{Effects of cell-size-aware optimistic upper-confidence bound.}
To investigate the importance of the component $L\ \mathrm{diam}(\cell)$ in the optimistic upper-confidence bound in \cref{eq:u_value}, \commNew{we} compare \solverAbbr and \solverAbbr(L=0). The results in \Cref{t:results_ablation} and \Cref{t:results_2} indicate that the cell-diameter-aware component in the upper-confidence bound in \cref{eq:u_value} is important for \solverAbbr to perform well, particularly in the SensorPlacement problems. The reason is that in the early stages of planning, the partitions associated to the beliefs are still coarse, \ie, only a few candidate actions are considered per belief. If some of those candidate actions have small estimated $Q$-values, \solverAbbr(L=0) may discard large portions of the action space for a very long time, even if they potentially contain useful actions. The cell-diameter-aware bias term in \cref{eq:u_value} alleviates this issue by encouraging \solverAbbr to explore cells with large diameters. This is particularly important for problems with large action spaces such as the SensorPlacement problems.

\paragraph{Effects of Stochastic Bellman backups.}
To investigate the effects of this component, let us compare \solverAbbr with \solverAbbr-MC. \Cref{t:results_ablation} and \Cref{t:results_2} reveal that \solverAbbr which uses stochastic Bellman backups often performs significantly better, particularly in the Parking2D and Parking3D problems. The reason is that in both problems good rewards are sparse, particularly for beliefs where the vehicle is located between the walls and slight deviations from the optimal actions can lead to collisions. The stochastic Bellman backups help to focus the search on more promising regions of the action space. \commNew{On the other hand, in the VDP-Tag problem, \solverAbbr-MC performs better than \solverAbbr. In this problem, it is important to reduce the uncertainty with respect to the target location during the first few steps by activating the range sensor. However, due to the cost of activating the range sensor, this strategy often seems suboptimal during the early stages of planning, when only a few episodes have been sampled. At the same time, in this problem, the stochastic Bellman backups in \solverAbbr tend to overestimate the $Q$-values for actions that do not activate the range sensor (this effect is known as \textit{maximization bias} in $Q$-learning\ccite{sutton2018reinforcement}). As a result, \solverAbbr tends to discard strategies that reduce the uncertainty with respect to the target location during the first few steps. \solverAbbr-MC which uses Monte-Carlo backups, does not suffer from maximization bias, which helps it to perform better than \solverAbbr in this problem.}

\subsubsection{Ablation Study}\label{sssec:abaltionStudy}

Here we investgate how \commNew{two} components of ADVT, \commNew{namely}, re-using partial search trees and using stochastic Bellman backups instead of Monte-Carlo backups, affect the baselines VOMCPOW and POMCPOW when we apply these ideas to the baselines. \commNew{The variants of VOMCPOW and POMCPOW that re-use partial search trees are denoted by VOMCPOW+\reuseTree\ and POMCPOW+\reuseTree\ respectively. The variants of the baselines that use stochastic Bellman backups are denoted by VOMCPOW+\bellmanBackup\ and POMCPOW+\bellmanBackup. The variants VOMCPOW+\reuseTree+\bellmanBackup\ and POMCPOW+\reuseTree+\bellmanBackup\ both re-use partial search trees and perform stochastic Bellman backups.}

\commNew{Generally, the results in \Cref{t:results_ablation} and \Cref{t:results_2} indicate that the baselines that re-use partial search trees perform much better than the baselines VOMCPOW and POMCPOW respectively, particularly in the Pushbox problems. These results (as well as those of \solverAbbr and all its variants) indicate the benefit of re-using partial search trees that were generated in previous planning steps instead of re-computing the policy at every planning step. Similarly, the baselines that use stochastic Bellman backups tend to outperform the ones that use Monte-Carlo backups, except for the VDP-Tag problem. This is consistent with our results for \solverAbbr and \solverAbbr-MC and indicates that stochastic Bellman backup is a simple, yet viable tool to improve the performance of MCTS-based solvers.}

\section{Conclusion}

We propose a new sampling-based online POMDP solver, called \solverAbbr, that scales well to POMDPs with high-dimensional continuous action \commNew{and observation} spaces. 
Our solver builds on a number of works that uses adaptive discretization of
the action space, and introduces a more effective adaptive discretization method
that uses novel ideas:
A Voronoi tree based adaptive hierarchical discretization of the action space, 
a novel cell-size aware refinement rule, 
and a cell-size aware upper-confidence bound. \commNew{For continuous observation spaces, our solver adopts the Progressive Widening and explicit belief representation strategy, enabling \solverAbbr to scale to higher-dimensional observation spaces.}
\solverAbbr shows strong empirical results against state-of-the-art algorithms
on several challenging benchmarks. We hope this work further expands the applicability of general-purpose POMDP solvers. 


\begin{acks}
We thank Jerzy Filar for many helpful discussions. 
This work is partially supported by the Australian Research Council (ARC)
Discovery Project 200101049. 
\end{acks}

\bibliographystyle{SageH}

\end{document}